\documentclass{article} 
\usepackage{iclr2025_conference,times}

\usepackage{caption}

\usepackage{amsmath,amsfonts,bm}

\def\eqref#1{equation~\ref{#1}}

\def\1{\bm{1}}

\DeclareMathAlphabet{\mathsfit}{\encodingdefault}{\sfdefault}{m}{sl}
\SetMathAlphabet{\mathsfit}{bold}{\encodingdefault}{\sfdefault}{bx}{n}

\usepackage{graphicx} 
\usepackage{booktabs}
\usepackage{soul}
\usepackage{gensymb}
\usepackage{listings}
\usepackage{rotating}

\usepackage{color}
\definecolor{orange}{RGB}{255,127,0}
\definecolor{brown}{RGB}{150,70,0}
\definecolor{green}{RGB}{127,255,127}
\definecolor{darkgreen}{RGB}{0,127,0}
\definecolor{blue}{RGB}{127,127,255}
\definecolor{lightblue}{RGB}{150,150,255}
\definecolor{darkblue}{RGB}{0,0,127}
\definecolor{red}{RGB}{255,90,90}
\definecolor{purple}{RGB}{200,110,170}
\definecolor{grey}{RGB}{127,127,127}
\definecolor{pink}{RGB}{255,180,180}

\newcommand{\sidenotes}{} 

\ifdefined\sidenotes

\newcommand{\sn}[1]{\textcolor{blue}{${\leftarrow}\hspace{-3pt}{\bullet}$\marginpar{\textcolor{blue}{${\leftarrow}\hspace{-3pt}{\bullet}$}\tiny{\textcolor{blue}{#1}}}}}

\newcommand{\snBen}[1]{\textcolor{purple}{${}\hspace{-3pt}{\bullet}$}\marginpar{\textcolor{purple}{${\leftarrow}\hspace{-3pt}{\bullet}$}\tiny{\textcolor{purple}{Ben: #1}}}}

\newcommand{\snMatteo}[1]{\textcolor{darkgreen}{${}\hspace{-3pt}{\bullet}$}\marginpar{\textcolor{darkgreen}{${\leftarrow}\hspace{-3pt}{\bullet}$}\tiny{\textcolor{darkgreen}{Matteo: #1}}}}

\newcommand{\snKozzy}[1]{\textcolor{red}{${}\hspace{-3pt}{\bullet}$}\marginpar{\textcolor{red}{${\leftarrow}\hspace{-3pt}{\bullet}$}\tiny{\textcolor{red}{Kozzy: #1}}}}

\newcommand{\snLucy}[1]{\textcolor{darkblue}{${}\hspace{-3pt}{\bullet}$}\marginpar{\textcolor{darkblue}{${\leftarrow}\hspace{-3pt}{\bullet}$}\tiny{\textcolor{darkblue}{Lucy: #1}}}}

\newcommand{\snMarko}[1]{\textcolor{brown}{${}\hspace{-3pt}{\bullet}$}\marginpar{\textcolor{brown}{${\leftarrow}\hspace{-3pt}{\bullet}$}\tiny{\textcolor{brown}{Marko: #1}}}}

\newcommand{\snJonathan}[1]{\textcolor{orange}{${}\hspace{-3pt}{\bullet}$}\marginpar{\textcolor{orange}{${\leftarrow}\hspace{-3pt}{\bullet}$}\tiny{\textcolor{orange}{Jonathan: #1}}}}

\newcommand{\todo}[1]{\textcolor{orange}{\sc To do: #1}} 
                
\else

\newcommand{\sn}[1]{}
\newcommand{\snBen}[1]{}
\newcommand{\snMatteo}[1]{}
\newcommand{\snKozzy}[1]{}
\newcommand{\snLucy}[1]{}
\newcommand{\snMarko}[1]{}
\newcommand{\snJonathan}[1]{}
\newcommand{\todo}[1]{}                
\fi

\title{A little less conversation, a little more action, please: Investigating the physical common-sense of LLMs in a 3D embodied environment}

\iclrfinalcopy 
\begin{document}



\author{
Matteo G. Mecattaf\thanks{Equal contribution, first author; emails: \texttt{mgmecattaf@gmail.com}, \texttt{bas58@cam.ac.uk}} \hspace{0mm} \thanks{Leverhulme Centre for the Future of Intelligence, University of Cambridge, Cambridge, UK} \And 
Ben Slater\footnotemark[1] \hspace{0mm} \thanks{Department of Psychology, University of Cambridge, Cambridge, UK} \hspace{0mm}  \footnotemark[2] \And
Marko Tešić\footnotemark[2] \And
Jonathan Prunty\footnotemark[2] \And
Konstantinos Voudouris\thanks{Equal contribution, senior author} \hspace{0mm} \thanks{Institute for Human Centered AI, Helmholtz Zentrum, Munich, Germany} \hspace{0mm} \footnotemark[2] \And
Lucy G. Cheke\footnotemark[4] \hspace{0mm} \footnotemark[3] \hspace{0mm} \footnotemark[2]
}

\maketitle

\begin{abstract}

As general-purpose tools, Large Language Models (LLMs) must often reason about everyday physical environments. In a question-and-answer capacity, understanding the interactions of physical objects may be necessary to give appropriate responses. Moreover, LLMs are increasingly used as reasoning engines in agentic systems, designing and controlling their action sequences. The vast majority of research has tackled this issue using static benchmarks, comprised of text or image-based questions about the physical world. However, these benchmarks do not capture the complexity and nuance of real-life physical processes. Here we advocate for a second, relatively unexplored, approach:~`embodying' the LLMs by granting them control of an agent within a 3D environment. We present the first embodied and cognitively meaningful evaluation of physical common-sense reasoning in LLMs. Our framework allows direct comparison of LLMs with other embodied agents, such as those based on Deep Reinforcement Learning, and human and non-human animals. We employ the Animal-AI (AAI) environment, a simulated 3D \textit{virtual laboratory}, to study physical common-sense reasoning in LLMs. For this, we use the AAI Testbed, a suite of experiments that replicate laboratory studies with non-human animals, to study physical reasoning capabilities including distance estimation, tracking out-of-sight objects, and tool use. We demonstrate that state-of-the-art multi-modal models with no finetuning can complete this style of task, allowing meaningful comparison to the entrants of the 2019 Animal-AI Olympics competition and to human children. Our results show that LLMs are currently outperformed by human children on these tasks. We argue that this approach allows the study of physical reasoning using ecologically valid experiments drawn directly from cognitive science, improving the predictability and reliability of LLMs.

\end{abstract}

\section{Introduction}

Large Language Models (LLMs) can do your physics homework, but might not be able to successfully find their way to the classroom. While LLMs have made great strides in several areas, including writing code \citep{champa2024chatgpt}, solving maths problems \citep{frieder2024mathematical, yuan2023well}, and answering general knowledge questions \citep{wang2024evaluating}, it remains unclear to what extent they can be considered to \textit{know} about and understand the physical world.


Physical common-sense reasoning is the capacity to perceive, understand, and predict the behaviour of objects in an environment. This includes an understanding of the physical rules governing space and objects in that environment, and how they might interact to determine the outcome of events or actions. In cognitive science, physical common-sense reasoning is also referred to as \textit{intuitive} or \textit{folk physics} \citep{kubricht2017intuitive}. In LLMs, this capability has typically been evaluated using task- or image-based benchmarks involving short vignettes describing a physical scene, perhaps accompanied by an image if the model is multi-modal, with questions about the objects and their interactions \citep{buschoff2023have, bisk2020piqa, wang2023newton}. Benchmark scores are then aggregated to produce the final estimate of an LLM's capability. While this traditional approach has given us an insight into some aspects of physical reasoning, it misses much of the definitive features of physical \textit{common sense} reasoning - that is, the capacity to \textit{perceive, understand, and predict} the behaviour of objects in a physical environment, and use their knowledge to take appropriate actions.

Beyond this, traditional benchmarks suffer from a number of shortcomings \citep{hernandez2017evaluation}. First, these benchmarks lack ecological validity---when deployed, LLM agents will not be interacting with well-described, clean vignettes with clear questions and uniquely identifiable answers. Instead, they will be interacting with a complex, noisy world where the correct answer, or action, is not always easily discriminated. Second, these benchmarks lack established construct validity \citep{borsboom2004concept,cronbach1955construct}---they have not been validated independently as \textit{good} measures of physical common-sense reasoning by, for example, running experiments with humans or animals. Third, these benchmarks are static, meaning that the test items are fixed. When these benchmarks are released, there is a risk that new models will be trained on test items, contaminating the benchmark and thus rendering any results invalid, since models have been trained to predict the answer rather than to exhibit any emergent physical common-sense reasoning \citep{xu2024benchmark}. Finally, benchmarks of physical common-sense reasoning are large and general---it is often unclear which \textit{aspects} of physical common-sense reasoning they are targeting for evaluation. This is problematic because this type of reasoning is multifaceted, comprising everything from understanding inertia, gravity, and the solidity of objects, to reasoning about the concepts of causality, quantity and time \citep{lake2017building,shanahan2020artificial}. Traditional benchmarks do not allow us to precisely answer questions about what LLMs know about their physical environments \textit{and} how they use that knowledge to take actions in them.

In this paper, we introduce \textit{LLMs in Animal-AI} (LLM-AAI), a framework for conducting robust cognitive evaluations of the physical common-sense reasoning capabilities of LLM agents in a 3D virtual environment. Our framework allows us to test LLMs' physical common sense reasoning by embodying LLMs within Animal-AI---a \textit{virtual laboratory} environment designed for the development of systematic cognitive test batteries with a particular emphasis on physical common-sense reasoning \citep{voudouris2023animal}. Our approach situates LLMs in a realistic physical environment (ecologically valid), draws on testing materials that have been independently validated on humans and other animals (construct valid), capitalises on the variance of physical phenomena to produce difficult, dynamic tests (non-static), and tests a range of components of physical common-sense reasoning (precise evaluation target). A further strength of the LLM-AAI framework is that it facilitates comparison between human, animal and multiple types of artificial intelligence systems on directly comparable tests. Here, we present the first evaluation of physical common-sense reasoning in LLMs using experiments drawn from research testing these capabilities in non-human animals, and compare their performance to Reinforcement Learning (RL) agents and human children. 

The paper proceeds as follows: First, we review the recent literature on LLM agents and physical common-sense reasoning evaluations. Second, we introduce the Animal-AI environment and the Animal-AI Olympics---a competitive cognitive benchmark drawing on experiments from comparative psychology. Third, we introduce the LLM-AAI framework and describe the results from two experiments, where we evaluate the performance of three state-of-the-art LLMs (Claude Sonnet 3.5, GPT-4o, and Gemini 1.5 Pro) on the Animal-AI Olympics, in comparison to RL agents and human children, using different prompting strategies. Finally, we discuss these results and future work developing the LLM-AAI framework.

\section{Related Work}

In machine learning and natural language processing, there has been increasing interest in whether Large Language Models possess the capacity to perceive, understand, and predict the behaviour of objects in their environment, which has come to be known in the literature as \textit{physical common-sense reasoning} (\citealt{bisk2020piqa,buschoff2023have,sap2020commonsense,storks2019commonsense,wang2023newton}; see also `world models', e.g., \citealt{matsuo2022deep}). 
This capacity has been studied extensively in the cognitive sciences, where it is often called \textit{intuitive} or \textit{folk physics} \citep{bates2019modeling, battaglia2012computational, chiandetti2011intuitive, povinelli2003folk, smith2018different}. Physical common-sense reasoning is multifaceted, ranging from understanding the properties and affordances of objects \citep{rutar2024general} to tracking occluded objects \citep{voudouris2022evaluating,voudouris2024investigating}, using tools \citep{shanahan2020artificial}, and predicting the effects of gravity and momentum \citep{buschoff2023have, jassim2023grasp, povinelli2003folk}. One approach to studying physical common-sense reasoning in Large Language Models is through the administration of text-based descriptions of physical scenes, sometimes accompanied by images in the case of multi-modal LLMs, about which the model must answer some questions. The \textit{Physical Interaction: Question Answering} (PIQA) benchmark \citep{bisk2020piqa} is a well-known benchmark of over 16K items that follows this approach, using only text-based questions. LLMs are asked how they might achieve certain goals, such as \textit{Make an outdoor pillow} and they are given two potential solutions, in this case, \textit{Blow into a trash bag and tie with a rubber band} or \textit{Blow into a tin can and tie with a rubber band}. Clearly, the answer is the former, given what we know as humans about the properties of trash bags and tin cans. \citet{aroca2021prost} extend PIQA with over 18K question-answer pairs in the PROST benchmark, and \citet{wang2023newton} scale up even further to over 160K items in the NEWTON benchmark. The results from these three benchmarks indicate that physical common-sense reasoning is not yet at human-level in LLMs. In the multi-modal context, \citet{buschoff2023have} develop a suite of tasks inspired by cognitive science to study physical common-sense among other things. In their design, multi-modal prompts including task descriptions and visual stimuli are combined, and LLMs are tasked with providing a numerical judgment or rating about the described physical scene. For example, in the \textit{block towers} task, LLMs are presented with pictures of stacks of coloured blocks, and asked to provide a binary judgment about whether the `tower blocks' are stable or not. In their results, they found that only OpenAI's GPT-4V was able to make correct judgments above the level of chance on this task. In a similar vein, \citet{jassim2023grasp} present the \textit{Grounding And Simulated Physics} (GRASP) benchmark, but in this case images are replaced with videos generated by a physics simulator. For every video, models are asked whether they think that the physical scene depicted is plausible, and they can only give a binary answer. Videos depict scenes in which objects appear to change size, colour, or shape spontaneously, disappear when occluded, or lack inertia or momentum. Their results also indicate that current LLMs that can process videos do not answer questions about these visual scenes above the level of chance.

An alternative approach to studying physical common-sense reasoning in LLMs is to grant them control of an agent, such that they are embodied in a real-world environment. Previous work has explored LLM embodiment via a number of different approaches in both physical and digital environments. In the field of robotics, LLMs have been used to generate high-level action plans that are executed in real-world settings \citep{ahn2022can,driess2023palm,jiang2022vima}. However for such forms of deployment to be safe and reliable, it is important to establish the extent to which LLM's impressive apparent understanding of the physical world translates into appropriate behaviour and decision-making when faced with real-world physical constraints \citep{ahn2022can}. Evaluating LLMs in `real-world' contexts offers a high degree of ecological validity, but presents significant challenges:~these approaches require extensive additional training, and face bottlenecks related to cost, safety and development speed in robotics. Hence, there is much to be gained from taking incremental steps towards true embodiment. One such step involves embedding LLMs as agents within virtual environments. While our focus is on physically realistic video games, there has also been work on using LLMs as Graphical User Interface (GUI) agents \citep{zhang2023mobile} or online assistants \citep{wang2024survey}.

While there has been considerable recent progress towards embodied LLM agents, there has been no work, to our knowledge, on providing a robust framework for evaluating their physical common-sense reasoning. In the remainder of this section, we briefly review research on LLM agents before comparing it to our approach. LLM agents have been implemented and evaluated in a wide variety of game environments \citep{hu2024survey}, ranging from co-operative games like \textit{OverCooked} \citep{agashe2023evaluating, gong2023mindagent, liu2023llm, zhang2023proagent} to strategy games like \textit{StarCraft II} \citep{ma2023large,shao2024swarmbrain}. Many of these games do not directly require good physical common-sense, because they involve simplistic visual and physical scenes with limited action spaces---their focus tends to be on evaluating how LLMs interact with other agents. In open field environments, there have been implementations of LLMs in  Minecraft \citep{chen2024s,fan2022minedojo,feng2023llama,liu2023llm,stengel2024regal,wang2023describe,wang2023jarvis,wang2023voyager,yuan2023skill,zhang2023creative,zhao2024hierarchical,zhu2023ghost} and Crafter \citep{du2023guiding,wu2024spring,zhang2023omni,zhang2024adarefiner}, although again the physical reality of these environments is heavily limited by their simplicity - indeed, Crafter is a 2D world \citep{hafner2021benchmarking}. Most closely aligned to our work are those LLM implementations in VirtualHome \citep{huang2022language,xiang2024language}, which has a realistic physics engine \citep{puig2018virtualhome}. In all cases, however, the focus has been on developing LLMs that can outperform humans or other agents, such as deep reinforcement learners, rather than developing a framework for better evaluation of physical common-sense reasoning. 

This paper is the first example of a novel framework and proof-of-concept results demonstrating that LLMs can be evaluated on ecologically valid, complex tasks of physical common-sense reasoning. Furthermore, our approach allows meaningful direct comparisons to be drawn between LLMs and other agents, both biological (e.g. children) and non-biological (e.g. Reinforcement Learning agents). This work is also part of a broader research effort that draws on methods from cognitive science and psychology to encourage greater predictive validity in AI evaluation, shifting the focus away from task-based benchmarks and leaderboards, toward broader capability-oriented evaluation \citep{burden2023inferring,burden2024evaluating,burnell2022not,hernandez2017evaluation}.

\section{The Animal-AI Environment}

The Animal-AI (AAI) environment \citep{beyret2019animalai, crosby2019animal, voudouris2023animal} is a 3D simulation based on the Unity ML-Agents framework \citep{juliani2018unity}, designed to be used by researchers from AI and cognitive science to assess nonverbal physical common sense reasoning in embodied agents. The goal of the environment is to offer a tool for interdisciplinary research at the intersection of AI and cognitive science, with a particular focus on comparative and developmental psychology. All experiments in AAI consist of a 40$\times$40 arena, populated with a single agent (spherical with diameter 1) and a variety of different objects.

\subsection{The Animal-AI Testbed and Olympics}

AAI was first released in 2019 as part of the Animal-AI Olympics Competition, in which over 60 entrants competed to produce agents that could solve a series of unseen tasks inspired by comparative psychology research \citep{crosby2020animal}, thus favouring the development of agents that could perform robustly \textit{out-of-distribution} on tests of physical common sense reasoning. After the competition was completed, these tasks were released as the Animal-AI Testbed to further stimulate interdisciplinary research between AI and comparative psychology. The Animal-AI Testbed contains 300 distinct tests (with 3 variants of each; n=900 tasks) that test the full breadth of capabilities that underpin physical common-sense reasoning, including navigating around obstacles, making spatial inferences, tracking occluded objects, and causal reasoning. The aim in every task is to maximise total reward at the end of the episode. The environment contains spheres of different colours and sizes: yellow spheres increase reward, as do green spheres, which also end the episode; red spheres decrease reward and end the episode. In all cases, the magnitude of the reward change is proportional to the size of the sphere. Touching red `death zones' leads to a decrease in reward of $-1$ and also ends the episode. Reward decreases at a constant rate starting from $0$ on each timestep, thus favouring efficient action sequences. Entering orange `hot zones' leads to a doubling in reward decrement. A variety of movable and immovable blocks are present in the environment, including tunnels and opaque and transparent walls. Ramps are always purple, platforms are always blue, and pushable blocks are always light grey. Other blocks may take any colour.

\begin{figure}[ht]
    \centering
    \includegraphics[width=1\linewidth]{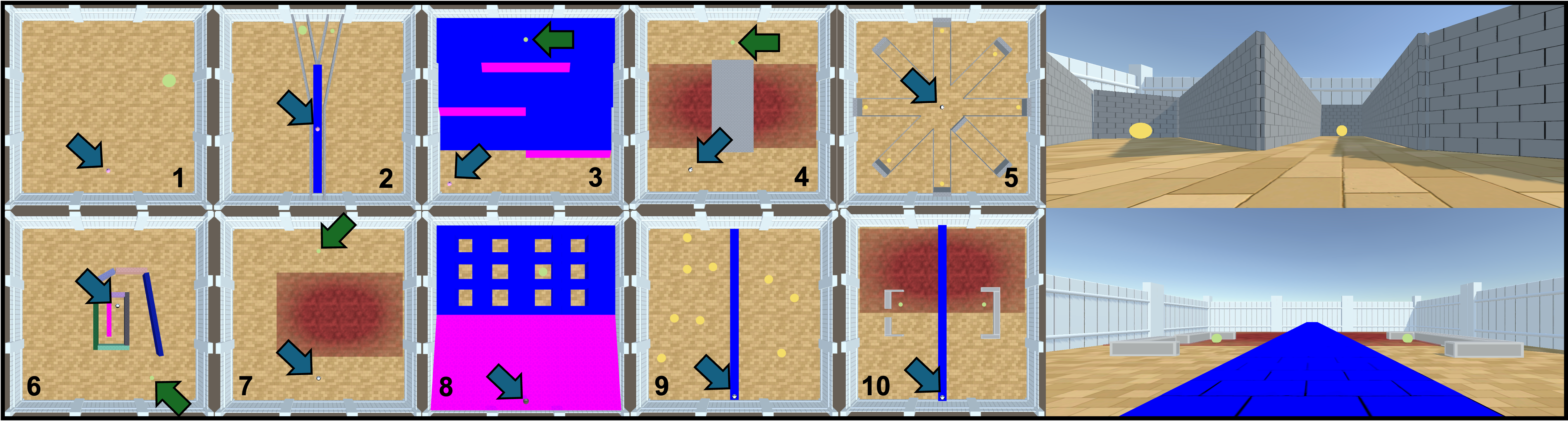}
    \caption{One task from each of the ten levels of the Animal-AI Testbed. The aim in every task is to collect as many yellow and/or green spheres while avoiding red zones, orange zones, and red spheres, before time runs out. Blue arrows indicate the location of the agent, and green arrows indicate the location of green spheres. The rightmost images show the agent's perspective during play in levels 5 and 10.}
    \label{fig:task-overview}
\end{figure}

The Animal-AI Testbed is arranged into 10 levels of 90 tasks of roughly increasing difficulty \citep{voudouris2022direct} which probe different aspects of physical common-sense reasoning. For example, level 1 (\textit{Food Retrieval}) tests the ability of the agent to navigate towards rewarding green and yellow spheres, level 2 (\textit{Preferences}) tests the ability to distinguish objects that give different rewards, and level 3 (\textit{Static Obstacles}) tests the ability to navigate around and over immovable solid objects, such as walls, ramps, and tunnels. The most complex levels test sophisticated physical common-sense reasoning abilities: level 8 (\textit{Object Permanence and Working Memory}) tests whether agents understand that objects continue to exist when they are occluded, while level 10 (\textit{Causal Reasoning}) tests the ability to understand cause and effect through the use of tools that can be used to achieve certain goals). These levels are described further in the Appendix in Table \ref{tab:olympics-test-bed}. Examples of the tests from each level used in this paper are presented in Figure \ref{fig:task-overview}.

\section{Methods}


\subsection{LLM-AAI}

\begin{figure}[ht]
    \centering
    \includegraphics[width=0.8\linewidth]{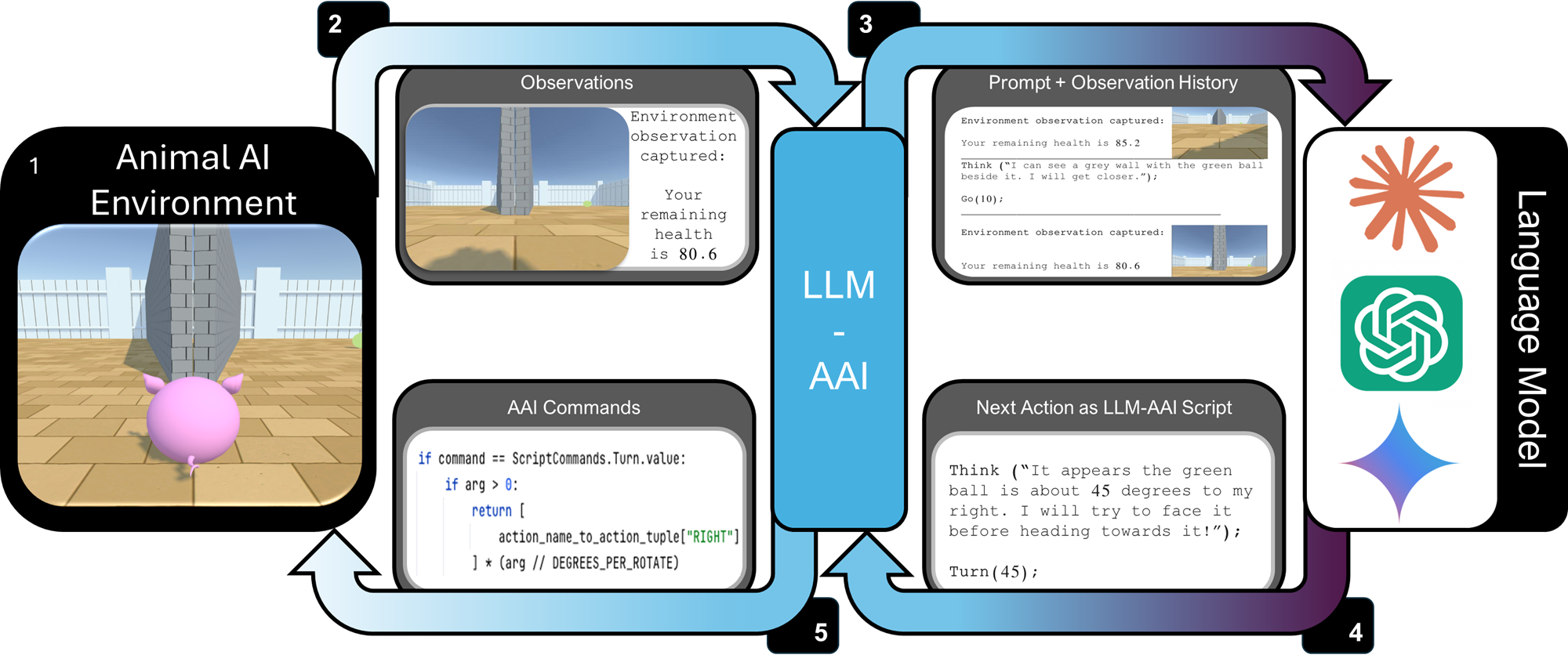}
    \caption{LLM-AAI. LLMs generate actions such as \texttt{Turn(45);} and pass them to LLM-AAI. LLM-AAI then parses these actions into commands that are understandable to the AAI environment and where they are subsequently executed. Observations from the environment are passed back to LLM-AAI, which concatenates them into the observation history and provides them, along with prompts like ``Your remaining health is 80.6'', to the LLM for reasoning and planning its next actions.}
    \label{fig:aai-llm}
\end{figure}

LLM-AAI framework allows us to connect LLMs with AAI environment. It is LLM-agnostic, requiring only a multimodal agent that can receive text-and-image inputs and return text outputs. Figure~\ref{fig:aai-llm} illustrates our approach. At each timestep, $t$, the environment returns a colour image of its current state, along with the agent’s current reward and health. These observations are combined into a prompt and presented to the LLMs as a request.

AAI requires an input on each frame describing how the agent should act (for example moving forwards or backwards, or rotating). We use an approach that finds a middle ground between requiring the LLM to provide such an input for each frame (which we discount for cost considerations), with approaches that require the LLM to interact with the environment by writing code that calls higher level APIs \citep{wang2023voyager} (which may outsource cognitively interesting tasks to specialised, environment-specific functions).
LLMs can act in the environment using a simple scripting language. The LLMs have access to three functions:
\begin{enumerate}
    \item \texttt{Go}---this command moves the agent forwards (positive integer) or backwards (negative integer). \texttt{Go(1);} moves the agent one unit forwards, where the units are in the size of the agent. Due to the momentum of moving objects in the environment, higher values take the agent slightly further than the number of units specified. For instance, crossing the width of the arena can be achieved with the \texttt{Go(35);} command, even though the arena is 40$\times$40 units.
    \item \texttt{Turn}---this command rotates the agent right (positive integer) or left (negative integer). The units are in degrees of arc. \texttt{Turn(-90);} rotates the agent 90\degree to its left, while \texttt{Turn(90);} rotates the agent 90\degree to its right. In AAI, the minimum amount of rotation is 6\degree, so all values in the \texttt{Turn} command are rounded down to the nearest multiple of 6.
    \item \texttt{Think}---the agent is instructed to use this command to describe the environment it observes, assess its position within that environment, track its remaining health and reward, and plan its course of action based on this information to collect the reward as efficiently as possible. For example, if the reward is behind the agent it might return \texttt{Think(`I think the reward is directly behind me: I will turn around to look for it');Turn(180);}. The inclusion of this command is influenced by approaches such as ReAct \citep{yao2022react}, in which LLM agents reason aloud.
\end{enumerate}

The LLM's response is parsed to return those scripts, which are converted into low-level action sequences, leading to a new state of the environment. Within a single episode, previous prompts and answers are prepended to the next prompt, so that the LLM has full access to previous states and action scripts. The LLM does not receive observations during the execution of action scripts.




\subsection{Large Language Models Tested}\label{sec:llms_tested}

We consider three state-of-the-art multi-modal Large Language Models. Our selection was based on a convenience sample, guided by the inclusion criterion that models must be multi-modal with a large context window ($>$64k), and the exclusion criterion that models must not be too costly to run inference on. We evaluated \textbf{Claude 3.5 Sonnet}, \textbf{GPT-4o}, and \textbf{Gemini 1.5 Pro}. We ran all experiments with temperature 0, but noticed that model responses can vary nevertheless. Therefore, we ran three trials of each model on each task.


\subsection{Experiments}

In this study\footnote{For this study, we use AAI version 3.1.3}, we use a subset of the Animal-AI Testbed containing four randomly selected tasks from the ten levels (n=40), replicating the design of \citet{voudouris2022direct}, in which 59 children aged 6-10 completed the same subset of 40 tasks. This allows direct comparison of LLM agents with human children, and non-human entrants to the Animal-AI Olympics Competition \citep{crosby2020animal}.

We conduct two experiments to explore LLM performance in this setting. Our first experiment includes a prompt that simply explains the environment and possible actions to the LLM, and assesses three models on 40 AAI Testbed tasks. Our second experiment provides the LLM with a prompt containing an in-context example of the successful completion of a simple `tutorial' level. We then evaluate LLMs given this prompt on a subset of the 40 tasks used in Experiment 1.

When we encountered errors from API calls that persisted after three retries, we discarded the current trial data and relaunched that trial run.
\subsubsection{Experiment 1}

First, we designed a simple prompt that provides the core information needed to navigate and collect rewards in the AAI Testbed.To improve the LLM's decision-making process, we incorporated the ReAct (Reasoning and Acting) framework \citep{yao2022react} into our prompt design. The ReAct approach combines reasoning and acting by allowing the model to generate reasoning traces alongside actions, which has shown improved performance on agentic tasks \citep{yao2022react}. By integrating ReAct, we encourage the LLM to first reason about the environment---identifying visible objects and their spatial relationships relative to the agent---before producing action scripts.

Our prompt begins by setting the context:~the LLM is informed that it is a player in a game set in a square arena with a white fence, tasked with collecting green and yellow ball rewards as quickly and efficiently as possible using a basic scripting language. The prompt details the kinds of objects the LLM will encounter, their key properties, and instructions on how to write scripts using the commands \texttt{Think}, \texttt{Go}, and \texttt{Turn}. It includes examples to illustrate correct usage of these commands and provides guidelines to avoid common mistakes.

To aid the LLMs in navigating the environment efficiently, we incorporated expert tips on movement distances and turning angles. For instance, we explain that moves of 1 to 10 steps cover small distances, while moves of 10 to 20 steps cover larger distances. We also provide strategic guidance on how to approach the task using the \texttt{Think} command to describe the current state of the environment and plan its actions, and subsequently using either \texttt{Go} or \texttt{Turn} to move within the environment. 

Lastly, the prompt warns about potential obstacles such as red lava puddles, holes, blue paths, purple ramps, transparent walls, pushable grey blocks, and immovable objects like walls and arches. It provides instructions on how to identify and interact with these obstacles, emphasizing caution to prevent the agent from dying or becoming trapped. The full prompt is provided in Appendix \ref{app:react-prompt}.



Armed with this prompt, each LLM is evaluated on the 40 tasks performed by children in Voudouris et al. \citep{voudouris2022direct}. The LLM is not presented with previous action scripts from other episodes, meaning it approaches each task as if it is interacting with the AAI Testbed for the first time.

\subsubsection{Experiment 2: Supervised In-Context Learning}


When children played the tasks in the AAI Testbed, they received a short two-minute video to describe ``the game''---that is, to introduce the AAI environment, its objects and controls. To emulate this, we designed an example level in AAI that introduced the same information as was presented in the video, and a sequence of scripts that could be used to solve the level, using the `Think' action to explain observations. The script and observations were incorporated into the prompt designed above. In this way, the LLMs are provided with images of objects they may encounter in a level, as opposed to just textual descriptions, and an `expert example' (shown in Appendix \ref{app:ICL-prompt}), before they are tasked with controlling the agent. We call this \textit{supervised in-context learning}.


Due to the increased cost of passing several images and a large amount of text for every episode, we conducted this experiment on a subset of the tasks. After carrying out Experiment 1 and observing close to zero performance in the later levels, we decided to focus on the first three levels of the AAI Testbed. These levels were designed as the simplest tasks and showed an expected decline in LLM performance from Level 1 to Level 3. Focusing on these initial levels provided a better opportunity to observe differences in performance, whereas the later levels, due to their difficulty, may have resulted in floor effects.

\section{Results}\label{sec:results}

\subsection{Experiment 1}

\begin{figure}[h!]
    \centering
    \includegraphics[width=1\linewidth]{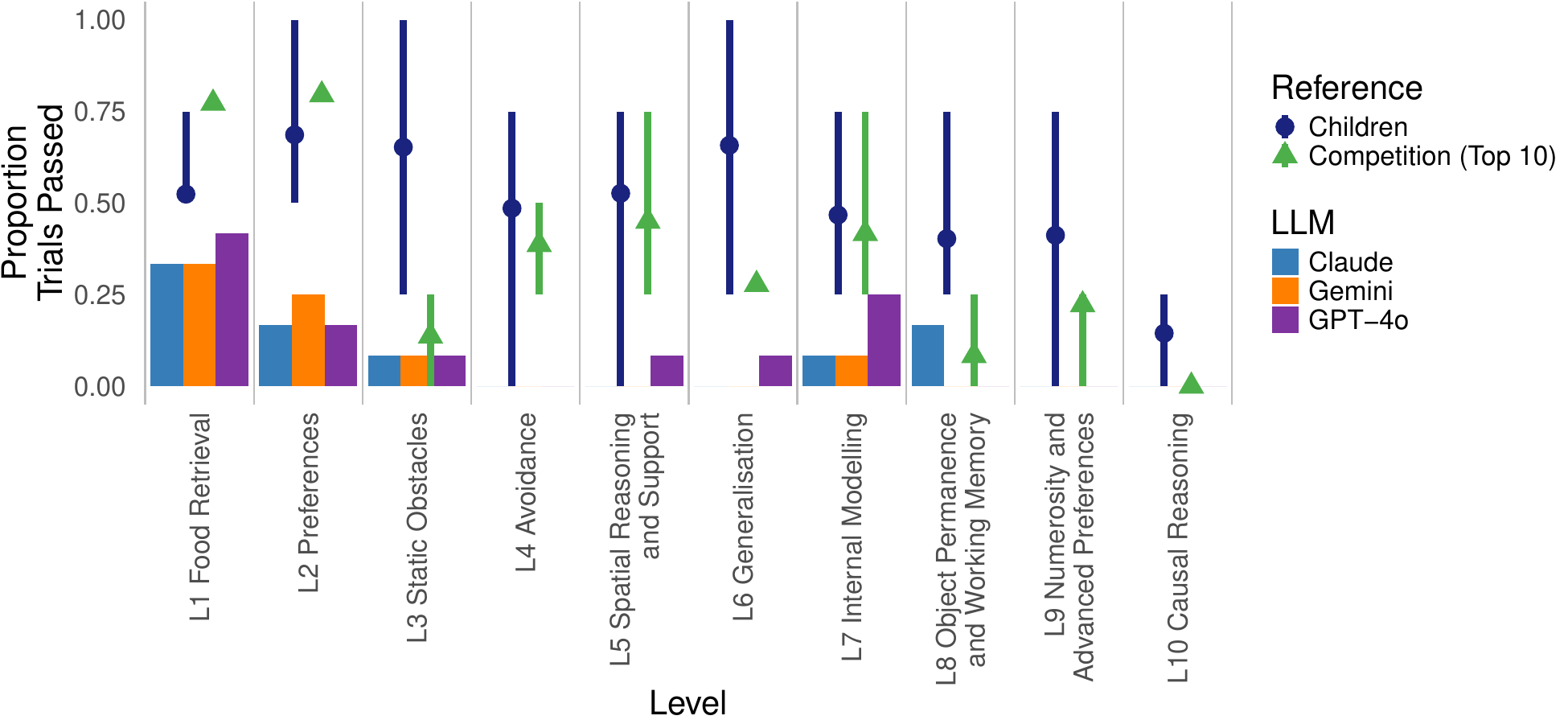}
    \caption{The proportion of trials passed by each LLM on each level, consisting of 3 trials of 4 tasks each (total n=12 trials per level). The interquartile range of proportions for all children (n=59) and the top 10 entrants to the Animal-AI Olympics Competition are presented as bars, with overall proportion for those populations indicated by points. Note that the children and competition agents have error bars, while the LLMs do not. This is because the child and competition agents contain a population of different individuals, across which we would like to understand variation, while the LLMs are repetitions of the same individual, and so are aggregated into a single value.}
    \label{fig:exp1-results}
\end{figure}

Our results, summarised in Figure \ref{fig:exp1-results}, show that LLMs are able to complete some challenges in Levels 1 and 2, with sporadic performance in across Levels 5, 6 and 8. They are comparable in performance with competition agents in Levels 3, 8, 9 and 10, however these all occur at a very low success rate, so there may be a floor effect obscuring a difference in capability between the groups. The children perform convincingly better than the LLM agents across all levels, with child error bars only overlapping with LLM performance in Levels 4, 5, 9 and 10, where LLM performance is very low.

These results show that LLMs are able to perform successfully in the most simple tasks of the testbed, but that their performance drops of quickly in more challenging tasks. The LLMs' performance never exceeds that of the top 10 agents submitted to the Animal-AI competition. It could be argued that this comparison will always favour the RL agents, who had been specifically trained for the environment, if not for the specific tasks. However, the same cannot be said for the human children, whose performance also exceeded that of the LLMs across the board. These results indicate that LLMs may still lack physical common-sense reasoning abilities possessed by human children.


\subsection{Experiment 2}
\begin{figure}[h!]
    \centering
    \includegraphics[width=1\linewidth]{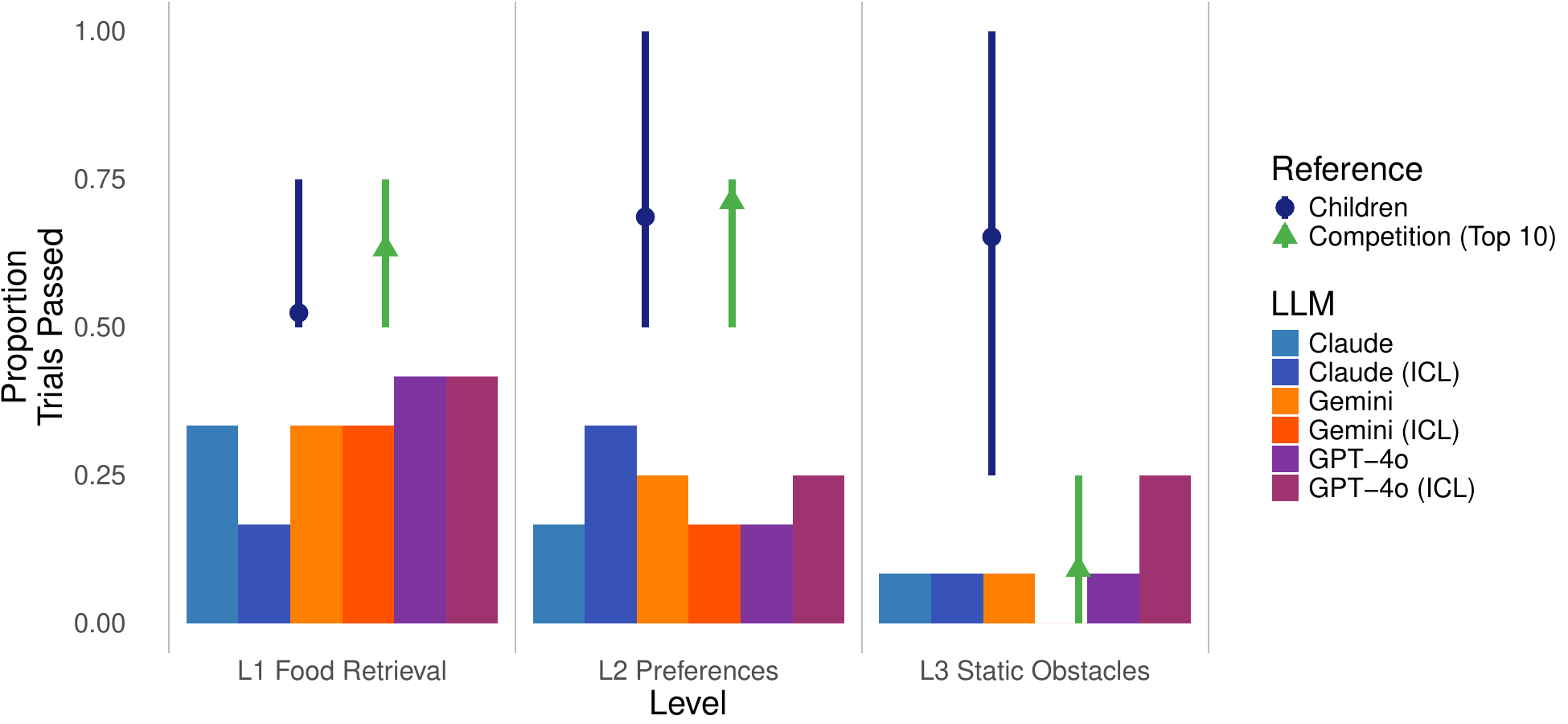}
    \caption{The proportion of trials by each LLM on each level, consisting of 3 trials of 4 tasks each (total n=12 trials per level). The interquartile range of proportions for all children
    (n = 59) and the top 10 entrants to the Animal-AI Olympics Competition are presented as bars, with
    overall proportion for those populations indicated by points. 
    }
    \label{fig:exp2-results}
\end{figure}

The results for our supervised in-context learning tasks are shown in Figure \ref{fig:exp2-results}. The performance of every tested LLM is illustrated by a pair of bars. The first bar illustrates performance \textit{without} our `expert example', and is identical to the Experiment 1 results from Figure \ref{fig:exp1-results}, while the second bar represents performance \textit{with} our example and is new in Experiment 2.

Overall, we did not observe a notable difference in performance when providing the LLMs with the `expert example. While the LLMs still broadly perform successfully on these early levels, they do not outperform the competition agents or the children.

The observed performance difference, when including the `expert example', was not the same across all the tested LLMs. Claude performed slightly worse in Level 1 than it had without in-context learning, whereas the opposite occurred in Level 2. Performance on Level 3 stayed the same. For Gemini, the addition of in-context learning had either no effect, in Level 1, or decreased the proportion of trials passed, in Levels 2 and 3. While GPT also experienced no performance difference in Level 1, its results rose both in Levels 2 and 3, with its Level 3 proportion of trials passed matching the upper interquartile range of the competition agents and the lower range of the children.


\section{Discussion}


The LLM-AAI framework tests the \textit{out of the box} physical reasoning capabilities of LLMs by using the ReAct prompting method \citep{yao2022react}, allowing LLMs to percieve and interact with the Animal-AI environment. While previous work has explored the capabilities of LLMs to interact with virtual environments, none have used this to explicitly develop a framework for testing physical common-sense reasoning in LLMs. Our results show that LLMs can not only be assessed in this way, but that when this is done it allows meaningful comparisons to be made with other biological and non-biological intelligences.

Evaluations in LLM-AAI have synergies with other efforts in evaluating and training LLMs. In evaluation, several LLM testbeds can be seen as targeting facets of the Animal-AI Testbed such as spatial reasoning \citep{ranasinghe2024learning}, numerosity \citep{trott2017interpretable, villa2023behind} and tool use \citep{tian2023macgyver}. Evaluations in LLM-AAI complement such efforts, adding the challenges of a 3D world, such as that of the complexity of 3D interactions, and that the target of the evaluation is less likely to be implied by the language of the prompt. Where a 3D environment has been used at the learning stage \citep{dagan2023learning, zellers2021piglet, driess2023palm, xiang2024language}, an LLM-AAI approach can be used to ensure the robustness of a model's physical common-sense.

For humans, an understanding of the physical world is built from countless embodied interactions with objects in their environment \citep{thelen2000grounded}. It is from these interactions that humans construct intuitive theories of the causal relationships that exist in their external world \citep{goddu2024development,gopnik2004mechanisms,tenenbaum2011grow}, and ground the symbolic concepts contained in language \citep{lakoff2008metaphors,wolff2007representing}. To date, there has been much debate as to the potential for `disembodied' systems such as LLMs to have a `meaningful' understanding of the physical world, or even a `world model' \citep{bender2020climbing,mitchell2021ai,shanahan2010embodiment}. The LLM-AAI framework allows us to make headway on these debates, with our initial results suggesting that LLMs still have some way to go before they can compete with their embodied counterparts.

\subsection{Limitations and Future Work}

The LLM-AAI framework satisfies an important demand in the field of Large Language Model evaluation. It provides a methodology and way forward for evaluations of physical common-sense reasoning using independently developed tests from cognitive science (construct valid) that measure specific components of physical common-sense (precise evaluation target), in a physically realistic environment (ecologically valid) with real-world dynamics (non-static). Furthermore, it enables direct, cognitively meaningful, comparisons between LLMs, deep reinforcement learning (DRL) agents, humans, and other animals. Our results in this paper demonstrate that out of the box systems can produce meaningful results on the Animal-AI competition. Nevertheless, there remain a number of extensions to how LLMs interact with AAI through our framework that could improve LLM performance. These extensions remedy some of the limitations of this current work and serve as the basis for future research.

\textbf{Sensing the environment.} In LLM-AAI, at every conversation turn, the tested LLM receives a single 512 x 512-pixel image of the environment. This image is captured after the LLM's action script is executed. The number of environment time-steps that unfold during the execution depends on the action script. For example, if the LLM uses the \texttt{Turn(180)} command, more environment time-steps will go by than if the LLM uses the \texttt{Turn(25)} command. Despite this difference in time-steps, in both cases a single image observation is sent to the LLM. While this observation routine allows larger agent-displacements with fewer API calls (and hence reduced costs), it can also cause the LLM to miss important environment information. For example, the agent may execute a \texttt{Turn(180)} script meaning that it misses the goal that is placed 90\degree to its right.



\textbf{Locomotion and control.} The control scheme used in the study, although theoretically sufficient for completing levels, is a relatively coarse way of controlling an agent in the environment compared to both children and AAI Olympics competition entrants, who could all provide a single action after every timestep. The additional challenge of writing action scripts manifests in the game-play of the LLMs. For example, in many cases, the LLM almost aligns itself with the goal but misses it slightly. This could result in the LLM finding itself beyond the goal and having to take extra turns to reorient itself before trying again. Future work could experiment with alternatives to the control scheme employed in this paper, such as allowing the LLM to control the agent frame-by-frame, or fine tuning a model to turn natural language descriptions of the action into environment commands.

\textbf{Capability limitations.} This study aimed to assess LLMs \textit{out of the box} on the Animal-AI Testbed. This has the benefit of ensuring that LLMs haven't been trained explicitly to solve these tests, thus contaminating the evaluation. However, it might be that the challenge of controlling the agent in the environment is so large that this dominates the cognitive challenge on some tasks. To address this, future work will fine-tune multi-modal LLMs on the observations and action scripts of an agent successfully completing simple navigation tasks. This would overcome the problem of calibrating action scripts to the environment, and allow our tests to more accurately reveal the cognitive capabilities of LLMs. An alternative approach would be to embed LLMs as components of a larger control and memory system \citep{wang2023voyager, sumers2023cognitive} to attempt achieve better performance on the Animal-AI Testbed.

\textbf{Cost.} The scaling cost of longer experiments rendered some experiments financially unfeasible. For example, human participants completing the same tasks as the LLM would have had the ability to learn over the course of the 40 arenas; this could be replicated in LLMs by attempting all 40 arenas in a single context window. However, the large number of tokens this generates is too costly. Due to financial limitations, the tested LLMs were also restricted to using, at most, 30 action-scripts, and therefore API calls, per episode. In contrast, human participants and DRL agents were only restricted by the arena's time-limit, rather than a maximum number of executed actions. This constraint was especially penalising for LLMs in arenas with many goals to find and in those requiring many fine movement and adjustments; such sequences inflated the number of action-scripts needed to complete the level. Future work will increase or remove the action-script limit and assess the change in performance.



\textbf{Towards cognitively-driven evaluation.} The levels in the Animal-AI Testbed are inspired by the rich tradition of developing non-verbal tests of capacities in cognitive science. Since there exists a large number of tests and experimental paradigms, they cannot be condensed into a single testbed such as ours. More targeted LLM-AAI evaluations using the tests from \citet{voudouris2022evaluating} for object permanence or \citet{rutar2024general} for object affordances, will allow assessors to make more precise statements the physical common-sense reasoning capabilities in this setting, and produce comparisons with the humans and DRL agents that have been evaluated on these tests.

\section{Conclusion}

We have introduced LLM-AAI, a framework for evaluating the physical common-sense reasoning capabilities of LLMs in a 3D environment. Using the diverse tasks of the Animal-AI Testbed, we have presented results from an initial assessment, showing that LLMs are capable of completing tasks using LLM-AAI, but may lack the physical common-sense reasoning capabilities of humans. We hope that these results will inspire researchers to embrace embodied evaluations as a powerful addition to the LLM evaluation toolbox.

\section{Reproducibility Statement}
All the results presented in this paper can be reproduced, provided that the closed-source LLM checkpoints that were tested are not altered. The checkpoints used were:
\begin{itemize}
    \item Claude 3.5 Sonnet:~claude-3-5-sonnet-20240620
    \item GPT-4o:~gpt-4o-2024-05-13
    \item Gemini 1.5 Pro:~gemini-1.5-pro-001
\end{itemize}

During our experiments we encountered issues with the API for Gemini 1.5 Pro, these issues were the only occasions in which we had to discard and rerun trials, as it stopped us from collecting complete data for trials. The API issue we encountered is documented at https://github.com/google-gemini/generative-ai-python/issues/559.

We also make the prompts that were passed to the LLMs available in Appendices \ref{app:react-prompt} and \ref{app:ICL-prompt}. We produced all of our results using Animal-AI version 3.1.3. Source code for our experiments is available at https://github.com/Kinds-of-Intelligence-CFI/LLM-AAI.


\section{Ethics Statement}

No human or animal participants were involved in this study, and no sensitive topics were used or contained in our interactions with LLMs. The human data used in our comparison was from an openly available dataset from an independent study found here: https://osf.io/g8u26/.

\section{Acknowledgments}

This work was partly funded under the Kinds of Intelligence project, The Leverhulme Centre for the Future of Intelligence (RC-2015-067), and an ESRC scholarship to BS (ES/P000738/1).

This research project used funds from the Microsoft Accelerate Foundation Models Research (AFMR) grant program to run experiments on GPT-4o.

\bibliography{bibliography}
\bibliographystyle{iclr2025_conference}
\newpage
\appendix

\section{The Animal-AI Testbed}

The Animal-AI Testbed contains 10 levels of 30 tasks with 3 variants each (n=900 tasks). Each level tests different aspects of physical common-sense reasoning. A description of each level is presented in Table \ref{tab:olympics-test-bed} overleaf. Participants in the Animal-AI Olympics Competition were tested on all 900 tasks of the Testbed, and developers were not given access to the contents of the Testbed prior to submission to the competition. In our plots in Section \ref{sec:results}, we only report the top 10 entrants to the competition in terms of overall score, indicating the current best performance of deep reinforcement learning (DRL) agents tested out-of-distribution. Data from children (n=59) on 4 tasks from each of the 10 levels (n=40) were taken from \citet{voudouris2022direct}. All comparisons between LLMs, children, and competition agents is based on their performances on only these 40 tasks.

\begin{sidewaystable}
\centering
 \caption{The Animal-AI Testbed consists of 10 levels of 30 tests with 3 variants each (n=900 tasks). Each level tests a different aspect of physical common sense reasoning.} 
  \begin{tabular}{p{0.4\linewidth} | p{0.6\linewidth}}
    \toprule
    Level     & Description  \\
    \midrule
    \rule{0pt}{1ex}
    L1 - Food Retrieval & Navigation towards rewarding objects in a large arena.\\
    \rule{0pt}{3ex}
    L2 - Preferences & Choice between objects with different reward values, indicated by their size and colour. \\
    \rule{0pt}{3ex}
    L3 - Static Obstacles & Objects partially occluded behind static obstacles around which agents must navigate. \\
    \rule{0pt}{3ex}
    L4 - Avoidance & Navigation around punishing objects to obtain rewarding objects. \\
    \rule{0pt}{3ex}
    L5 - Spatial Reasoning and Support & Based on the absence of rewarding objects in one part of the arena, agents must infer their presence elsewhere, even when (partially) occluded. Rewarding objects may also be out of reach on a ledge or a pillar, requiring the agent to push a movable block to knock them down. \\
    \rule{0pt}{3ex}
    L6 - Generalisation & Tasks from the first five levels are adapted so that the surroundings are colours. \\
    \rule{0pt}{3ex}
    L7 - Internal Modelling & Tasks from the first five levels with alternating periods in which visual information is withheld, as though the lights have gone out. The agent must continue to model their environment during these periods. \\
    \rule{0pt}{3ex}
    L8 - Object Permanence and Working Memory & Rewards are hidden behind obstacles for the agent to find. \\
    \rule{0pt}{3ex}
    L9 - Numerosity and Advanced Preferences & Discrimination between different numbers of rewarding objects, testing the ability to visually discriminate and count objects in a scene. \\
    \rule{0pt}{3ex}
    L10 - Causal Reasoning & Rewarding objects are only obtainable using a tool with certain physical properties and affordances. \\
    \bottomrule
  \end{tabular}
  \label{tab:olympics-test-bed}
\end{sidewaystable}

\newpage
\section{Initial Prompt}\label{app:react-prompt}

\begin{lstlisting}
You are a PLAYER in a game set in a square arena with a white fence.  Your task is to collect all the rewards as quickly and efficiently as possible using a basic scripting language. The rewards are green and yellow balls.

To successfully collect a reward, you must fully pass through it. For example, if you think the reward is 10 steps away, you should go further than 10 steps to ensure you collect it, e.g., Go(15);.

The game ends when you have collected all the rewards and the arena closes. If you are still in the arena, the game is NOT finished and you have NOT collected all the rewards.

Your remaining health is displayed in the environment as "Your remaining health is:". The game will end if your health reaches 0.

NOTE: When you collect a reward, your remaining health will INCREASE compared to the previous timestep. If it doesn\'t increase, the reward was not collected. Always compare your current health with the previous timestep to confirm this. The scripting language consists of commands in the form <COMMAND>(<ARG>);

Note:
- If ARG is numerical it should always be an integer, never a float.
- DO NOT include any response not following the format of the scripting language. Doing so will result in failure.
- DO NOT wrap your commands with inverted commas: \' \'Think(\'Something\');\'Go(5);\' \' would fail whereas \' Think(\'Something\');Go(5); \' would not.

Commands are:

- Think: Reason about what actions to take to collect the rewards most efficiently (does not affect the environment). Note: Always format the thought as a string. Also, when using this command, do not include parentheses as arguments. For example, correct: \'Think(\'I cannot see the reward---yellow or green ball---in the arena\');\' Incorrect: \'Think(\'I cannot see the reward (yellow or green ball) in the arena\');\'

- Go: Move forward or backward a certain number of steps (1 to 35 steps forward, -1 to -35 backward).

- Turn: Turn by a specified number of degrees (any positive number between 1 and 360 degrees turns the character to the right (clockwise) and any negative number between -1 and -360 degrees turns the character to the left (anticlockwise)).

Examples:
To move forward by 5 steps: \'Go(5);\'.
To investigate what is happening to your left: \'Think(\'I would like to investigate what is happening to my left\');Turn(-90);\'

The number of scripts you can send is limited, so try to complete the levels efficiently.
The size of the arena is 35 by 35: \'Go(35)\' will take you from one end of the arena to the other.
After you submit your script, you will receive an image observation. Use this image to plan your next script.

EXPERT TIPS:
- Moves of 1 to 10 steps cover small distances, while moves of 10 to 20 cover larger distances.
- Turns of 25 to 45 degrees turn you a small amount to the right, while turns of -25 to -45 degrees will turn you a small amount to the left. DO NOT use turns less than 25 degrees.
- Turns of 45 to 90 degrees will turn you a large amount to the right, while turns of -45 to -90 degrees will turn you a large amount to the left.\n
- Turning 180 or -180 degrees will turn you all the way round so that you are facing backwards.

How to approach the task:

Start by using the \'Think\' command to describe the environment you see. When you find the rewards, i.e. green or yellow balls, ALWAYS explicitly state BOTH your DISTANCE and ANGLE with respect to them. Note: Only green and yellow balls are rewards and nothing else.
Take appropriate actions. Use \'Go\' OR \'Turn\', but DO NOT combine them in the same turn. Always follow \'Think\' with one of these two actions.

HINT: Your vision is good but not perfect and some rewards may not be immediately visible. Rewards may be behind you. Explore the arena to locate them. When exploring, try to get a 360-view of the arena. If both green and yellow balls are present, collect the yellow balls first and green balls last. Note that some arenas may not have green balls at all. The reward you get is proportional to the size of the ball: make sure to get the bigger balls first!. Finally, the lights may go out during a level. They may or may not come back on: use what you\'ve learnt about the arena so far to move around and collect the reward when this happens!

When you find a reward: 
Use the \'Turn\' command to align yourself directly with the reward. Before moving towards it, check the observation image provided by the environment to ensure the reward is centered in your view. If the reward is not centered, adjust your alignment with additional turns until it is.
Use the \'Go\' command to move toward the reward.
If the reward is more than 15 steps away, align yourself with the reward as best as you can and move half the distance first. Then reassess your angle with respect to the reward, use \'Turn\' to adjust your angle if the reward is not centered in your view, and move the remaining distance.
Remember: ALWAYS check your health after collecting a reward. You have successfully collected the reward only if your health has INCREASED compared to the previous timestep.

Be mindful of obstacles:

Red lava puddles and red balls: If you run into them, you will die.
Holes: Some may contain rewards, but if you fall into an empty hole, you will be trapped and unable to collect other rewards.
Blue paths: These are slightly raised paths. You can walk on them, but once you step off, you won't be able to get back onto them.
Purple ramps: You can climb them to get to the other side. Once you climb over the ramp, you cannot climb back over the same ramp.
Transparent walls: You can see through them, but you cannot walk through them.
Pushable grey blocks: These are cube-like structures, patterned with dark grey rectangles on each face. If viewed from one side, they will look like a rectangular structure. They can be pushed, but they are heavy! To move these blocks, you need to run into them. The blocks are heavy so you need to add extra steps to your Go() command.
Immovable objects: Walls and arches cannot be moved.
Ready to play? You will start by seeing three image observations.
A new level begins now. Environment observation captured
\end{lstlisting}
\label{sec:prompt_text}

\newpage
\section{ICL Prompt}\label{app:ICL-prompt}

In Experiment 2 the initial prompt was accompanied by a demonstration of an episode, which included examples of objects it may encounter in AAI. We replicate this below in human-readable format; with observations in sequence, and their responses below:


\begin{center}
  \begin{minipage}[t]{0.3\textwidth}
    \centering
    \includegraphics[width=\linewidth]{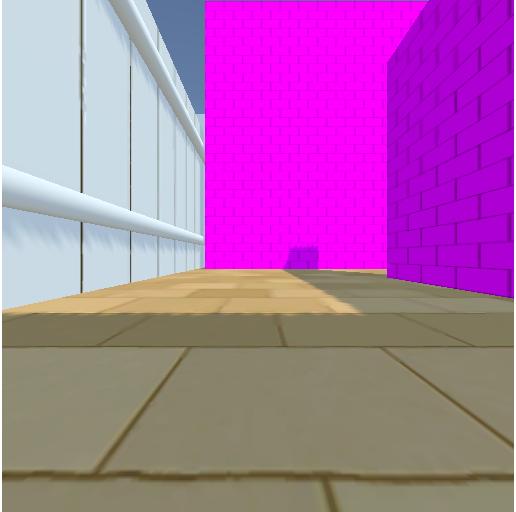}
    \captionof{figure}{$\langle$ Initial image: no response $\rangle$}
  \end{minipage}%
  \hfill
  \begin{minipage}[t]{0.3\textwidth}
    \centering
    \includegraphics[width=\linewidth]{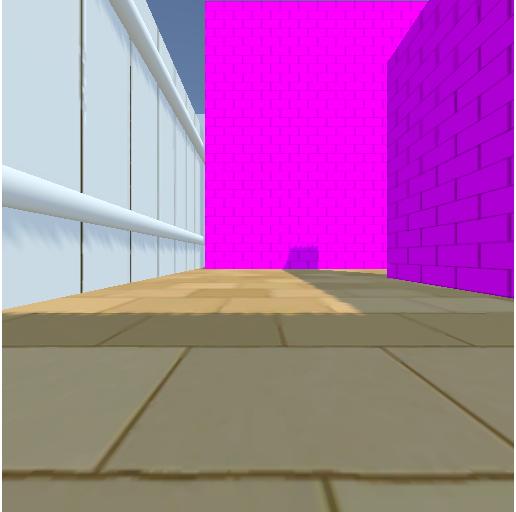}
    \captionof{figure}{$\langle$ Initial image: no response $\rangle$}
  \end{minipage}%
  \hfill
  \begin{minipage}[t]{0.3\textwidth}
    \centering
    \includegraphics[width=\linewidth]{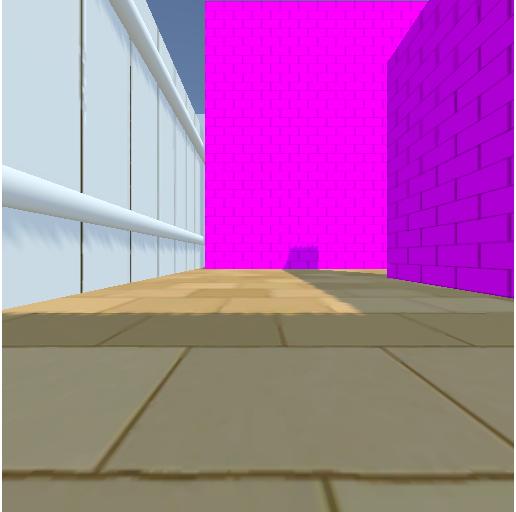}
    \captionof{figure}{Think(``I can move forwards and backwards with Go");Go(3);}
  \end{minipage}

  \begin{minipage}[t]{0.3\textwidth}
    \centering
    \includegraphics[width=\linewidth]{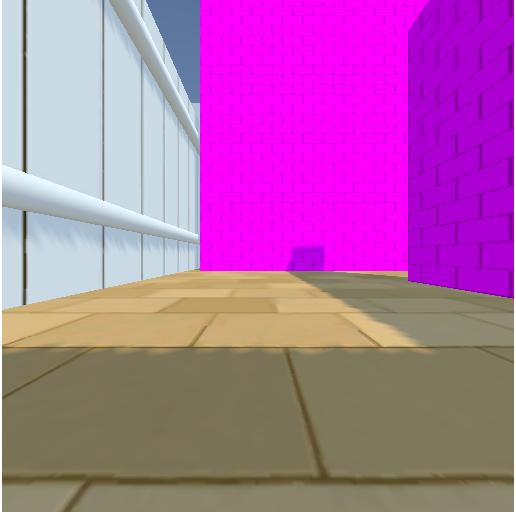}
    \captionof{figure}{Go(-3);}
  \end{minipage}%
  \hfill
  \begin{minipage}[t]{0.3\textwidth}
    \centering
    \includegraphics[width=\linewidth]{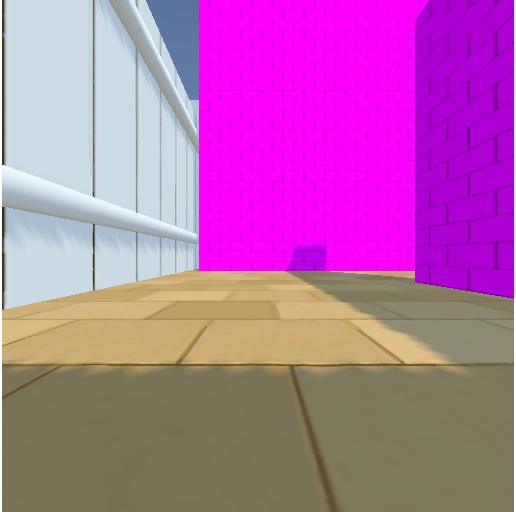}
    \captionof{figure}{Think(``I can turn left and right with Turn");Turn(30);}
  \end{minipage}%
  \hfill
  \begin{minipage}[t]{0.3\textwidth}
    \centering
    \includegraphics[width=\linewidth]{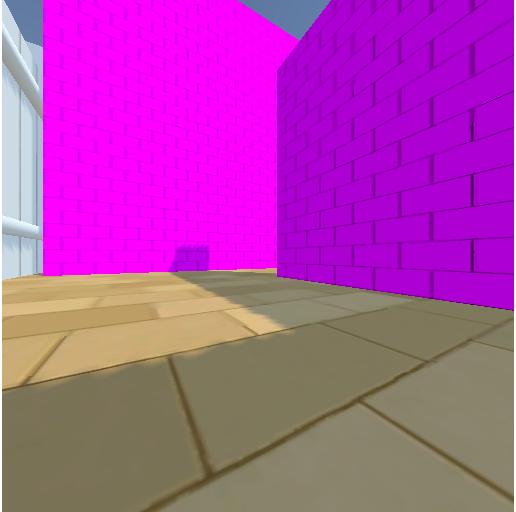}
    \captionof{figure}{Turn(-30);}
  \end{minipage}

  \begin{minipage}[t]{0.3\textwidth}
    \centering
    \includegraphics[width=\linewidth]{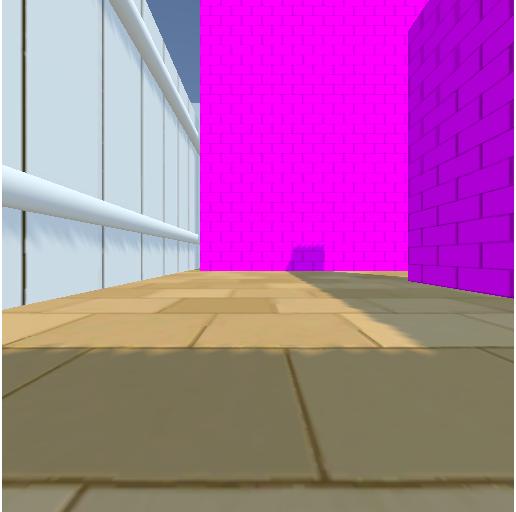}
    \captionof{figure}{Think(``I'll move through this level and take a look at some of the items I might encounter");Go(9);}
  \end{minipage}%
  \hfill
  \begin{minipage}[t]{0.3\textwidth}
    \centering
    \includegraphics[width=\linewidth]{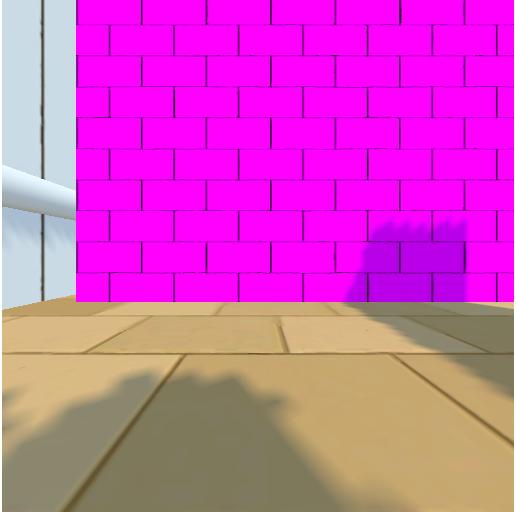}
    \captionof{figure}{Turn(90);}
  \end{minipage}%
  \hfill
  \begin{minipage}[t]{0.3\textwidth}
    \centering
    \includegraphics[width=\linewidth]{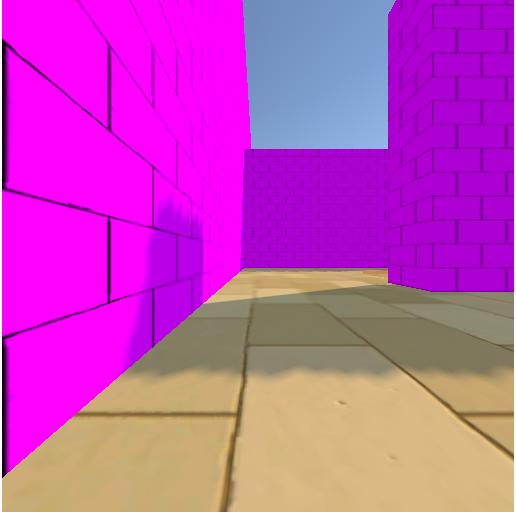}
    \captionof{figure}{Go(9);}
  \end{minipage}

  \begin{minipage}[t]{0.3\textwidth}
    \centering
    \includegraphics[width=\linewidth]{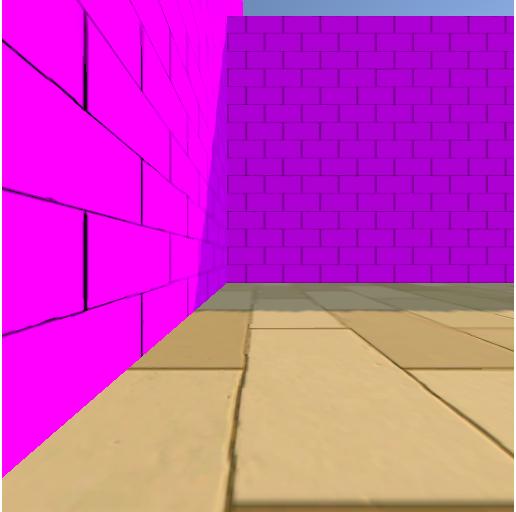}
    \captionof{figure}{Turn(90);}
  \end{minipage}%
  \hfill
  \begin{minipage}[t]{0.3\textwidth}
    \centering
    \includegraphics[width=\linewidth]{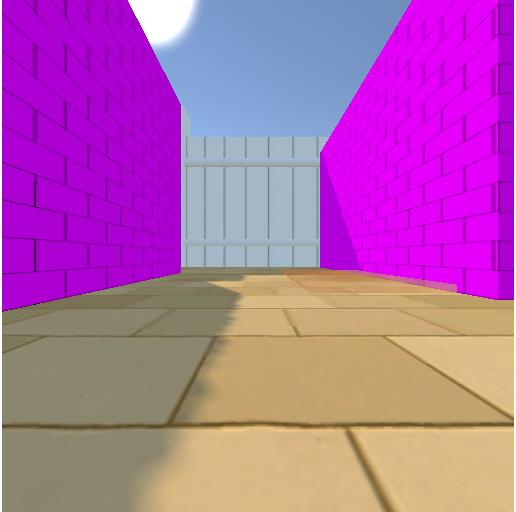}
    \captionof{figure}{Think(``I shouldn't stand on the orange zone for too long. I'll lose health more quickly!");Go(10);}
  \end{minipage}%
  \hfill
  \begin{minipage}[t]{0.3\textwidth}
    \centering
    \includegraphics[width=\linewidth]{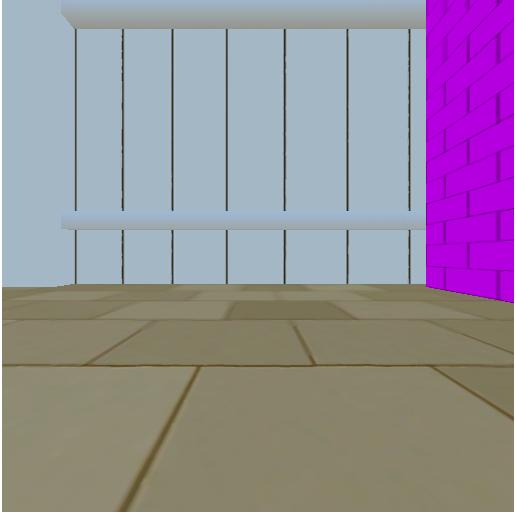}
    \captionof{figure}{Turn(-90);}
  \end{minipage}

  \begin{minipage}[t]{0.3\textwidth}
    \centering
    \includegraphics[width=\linewidth]{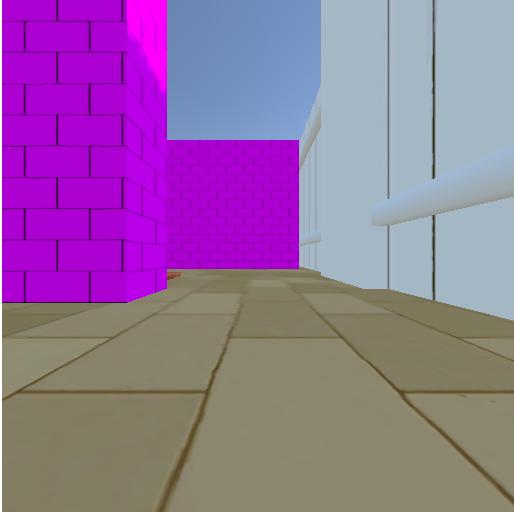}
    \captionof{figure}{Go(9);}
  \end{minipage}%
  \hfill
  \begin{minipage}[t]{0.3\textwidth}
    \centering
    \includegraphics[width=\linewidth]{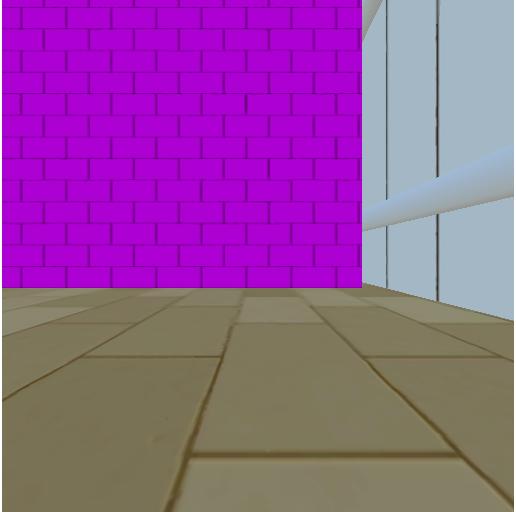}
    \captionof{figure}{Turn(-90);}
  \end{minipage}%
  \hfill
  \begin{minipage}[t]{0.3\textwidth}
    \centering
    \includegraphics[width=\linewidth]{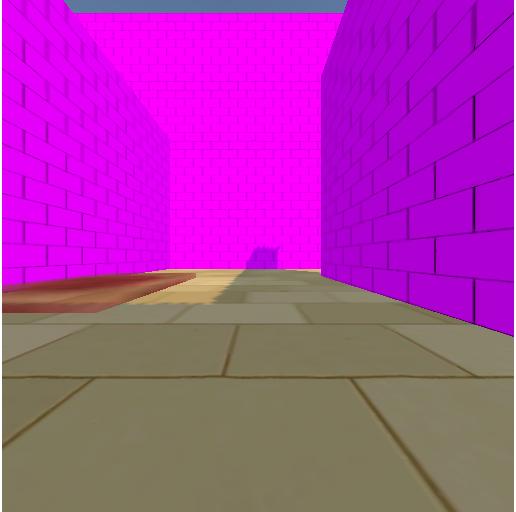}
    \captionof{figure}{Think(``I shouldn't step in the red zone at all: that will make me lose the level!");Go(9);}
  \end{minipage}

    \begin{minipage}[t]{0.3\textwidth}
    \centering
    \includegraphics[width=\linewidth]{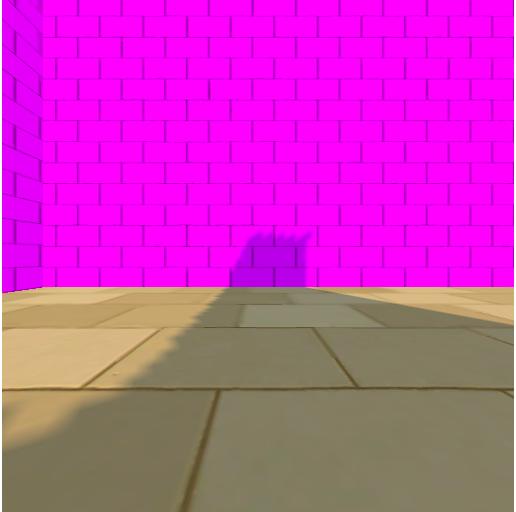}
    \captionof{figure}{Turn(90);}
  \end{minipage}%
  \hfill
  \begin{minipage}[t]{0.3\textwidth}
    \centering
    \includegraphics[width=\linewidth]{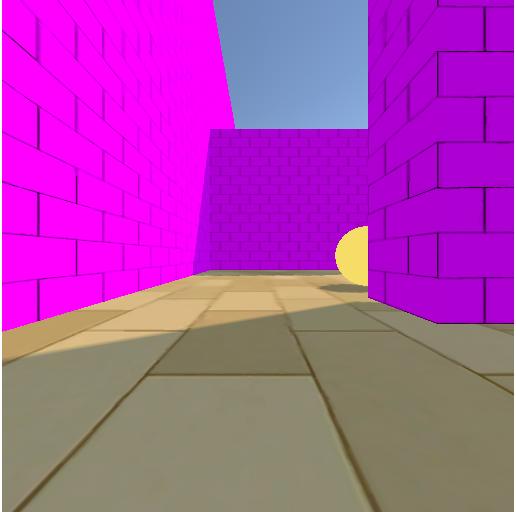}
    \captionof{figure}{Go(7);}
  \end{minipage}%
  \hfill
  \begin{minipage}[t]{0.3\textwidth}
    \centering
    \includegraphics[width=\linewidth]{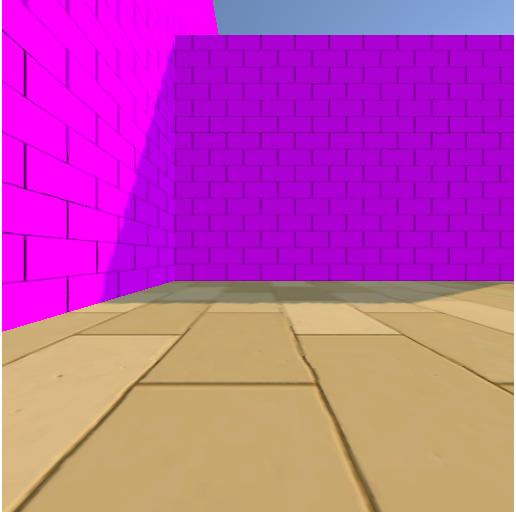}
    \captionof{figure}{Turn(90);}
  \end{minipage}

  \begin{minipage}[t]{0.3\textwidth}
    \centering
    \includegraphics[width=\linewidth]{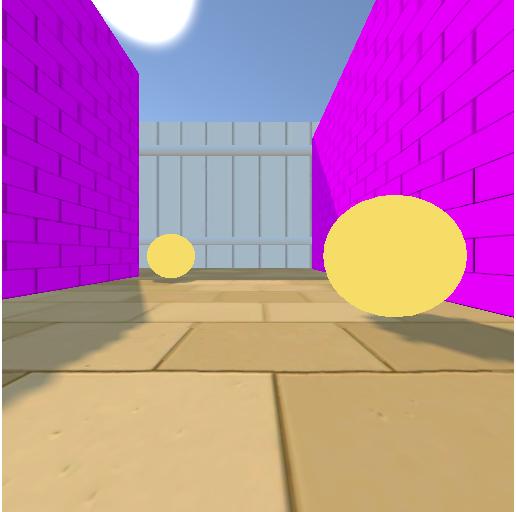}
    \captionof{figure}{Think(``I should collect as many yellow balls as possible, they'll increase my health!");Turn(30);}
  \end{minipage}%
  \hfill
  \begin{minipage}[t]{0.3\textwidth}
    \centering
    \includegraphics[width=\linewidth]{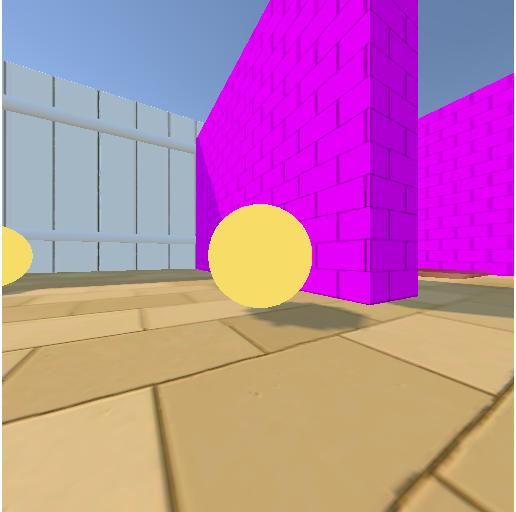}
    \captionof{figure}{Go(5);}
  \end{minipage}%
  \hfill
  \begin{minipage}[t]{0.3\textwidth}
    \centering
    \includegraphics[width=\linewidth]{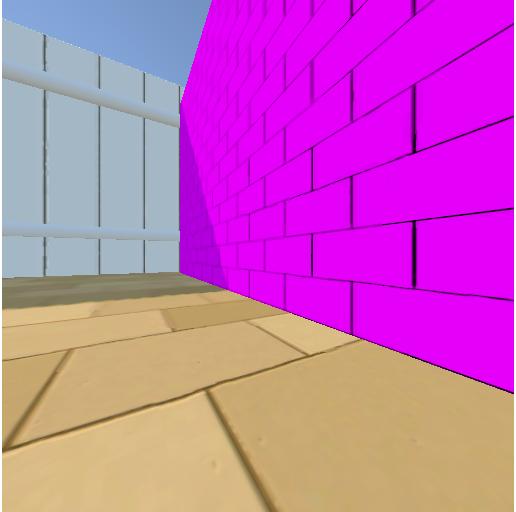}
    \captionof{figure}{Turn(-90);}
  \end{minipage}

  \begin{minipage}[t]{0.3\textwidth}
    \centering
    \includegraphics[width=\linewidth]{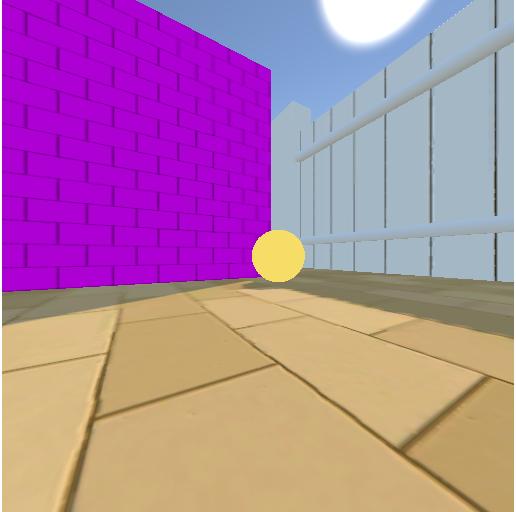}
    \captionof{figure}{Go(9);}
  \end{minipage}%
  \hfill
  \begin{minipage}[t]{0.3\textwidth}
    \centering
    \includegraphics[width=\linewidth]{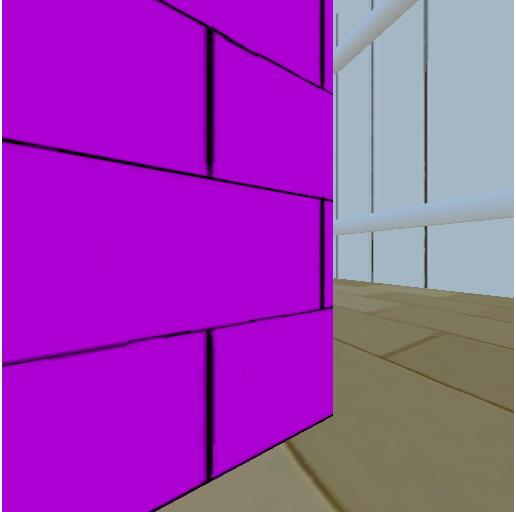}
    \captionof{figure}{Turn(45);}
  \end{minipage}%
  \hfill
  \begin{minipage}[t]{0.3\textwidth}
    \centering
    \includegraphics[width=\linewidth]{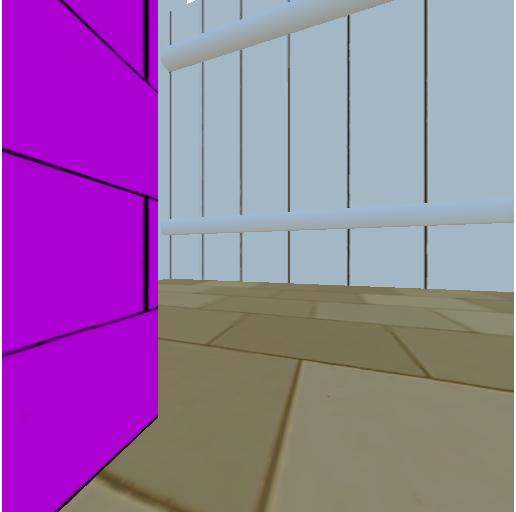}
    \captionof{figure}{Go(3);}
  \end{minipage}

    \begin{minipage}[t]{0.3\textwidth}
    \centering
    \includegraphics[width=\linewidth]{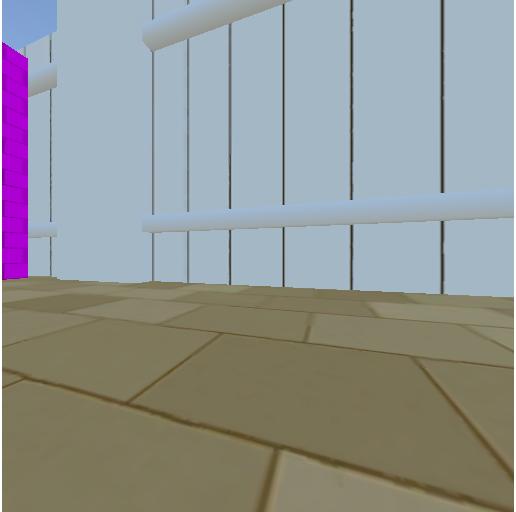}
    \captionof{figure}{Turn(-45);}
  \end{minipage}%
  \hfill
  \begin{minipage}[t]{0.3\textwidth}
    \centering
    \includegraphics[width=\linewidth]{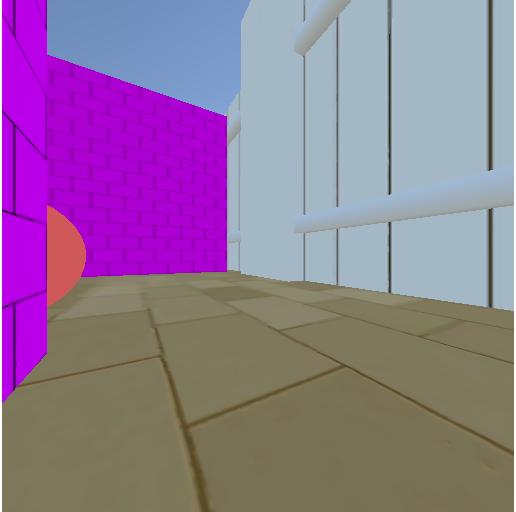}
    \captionof{figure}{Go(4);}
  \end{minipage}%
  \hfill
  \begin{minipage}[t]{0.3\textwidth}
    \centering
    \includegraphics[width=\linewidth]{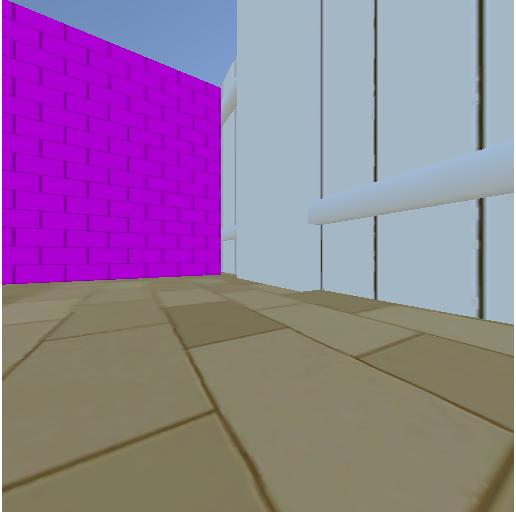}
    \captionof{figure}{Turn(-90);}
  \end{minipage}

    \begin{minipage}[t]{0.3\textwidth}
    \centering
    \includegraphics[width=\linewidth]{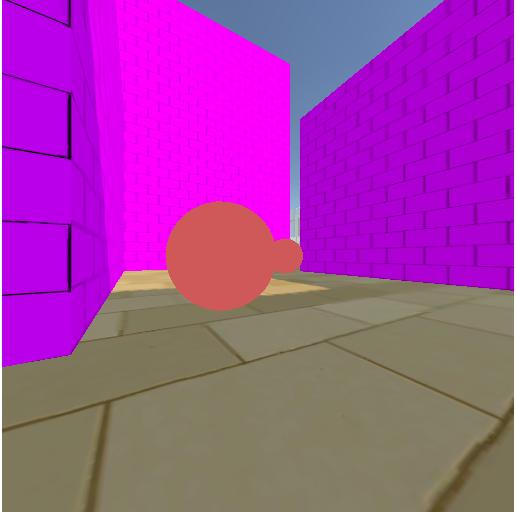}
    \captionof{figure}{Think(``I should be very careful to avoid the red balls: if I touch one of those I'll lose the level!");Turn(45);}
  \end{minipage}%
  \hfill
  \begin{minipage}[t]{0.3\textwidth}
    \centering
    \includegraphics[width=\linewidth]{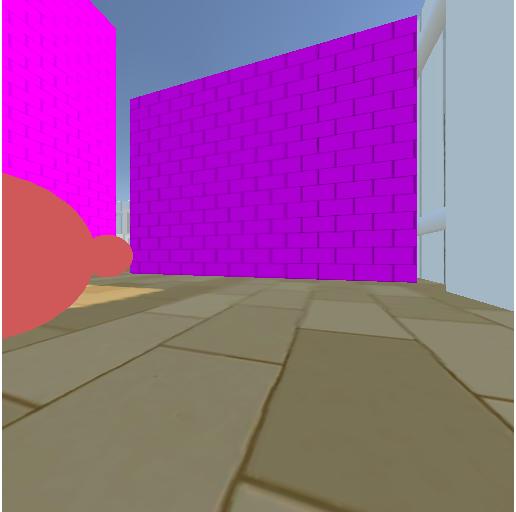}
    \captionof{figure}{Go(4);}
  \end{minipage}%
  \hfill
  \begin{minipage}[t]{0.3\textwidth}
    \centering
    \includegraphics[width=\linewidth]{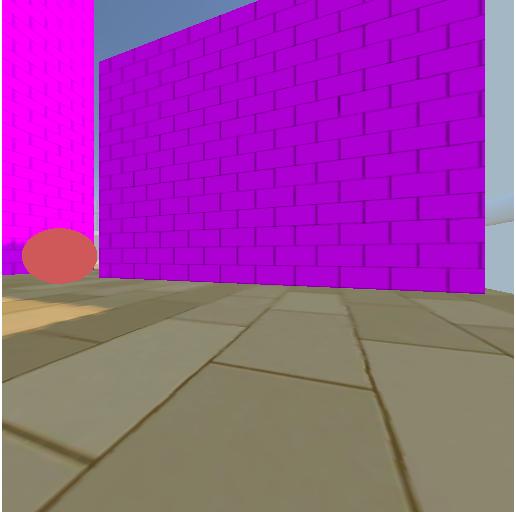}
    \captionof{figure}{Go(2);}
  \end{minipage}

    \begin{minipage}[t]{0.3\textwidth}
    \centering
    \includegraphics[width=\linewidth]{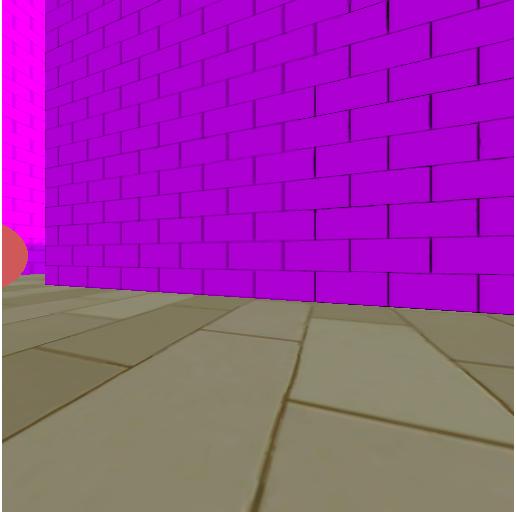}
    \captionof{figure}{Turn(-90);}
  \end{minipage}%
  \hfill
  \begin{minipage}[t]{0.3\textwidth}
    \centering
    \includegraphics[width=\linewidth]{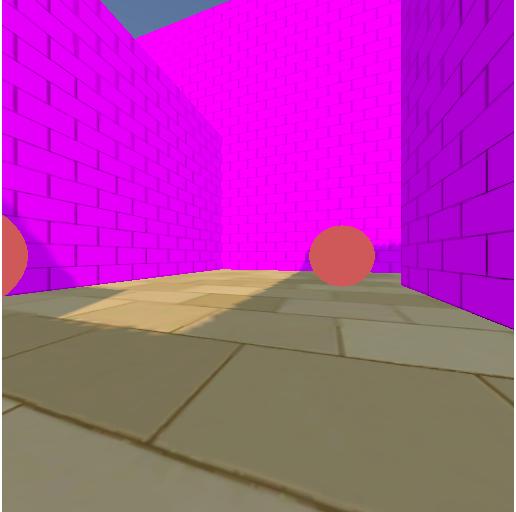}
    \captionof{figure}{Go(2);}
  \end{minipage}%
  \hfill
  \begin{minipage}[t]{0.3\textwidth}
    \centering
    \includegraphics[width=\linewidth]{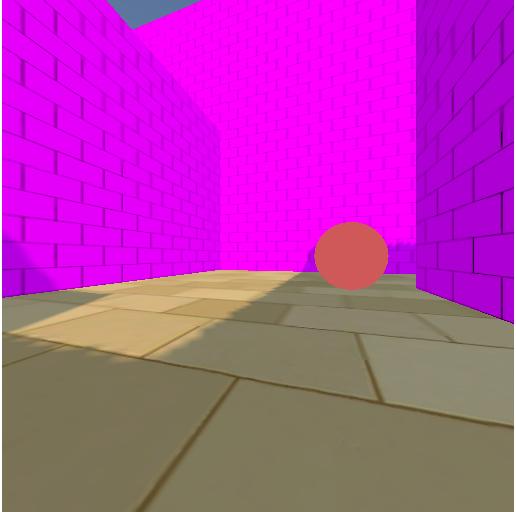}
    \captionof{figure}{Turn(-15);}
  \end{minipage}

    \begin{minipage}[t]{0.3\textwidth}
    \centering
    \includegraphics[width=\linewidth]{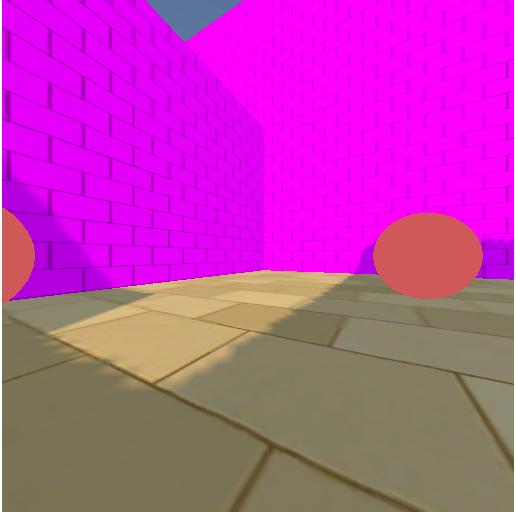}
    \captionof{figure}{Go(2);}
  \end{minipage}%
  \hfill
  \begin{minipage}[t]{0.3\textwidth}
    \centering
    \includegraphics[width=\linewidth]{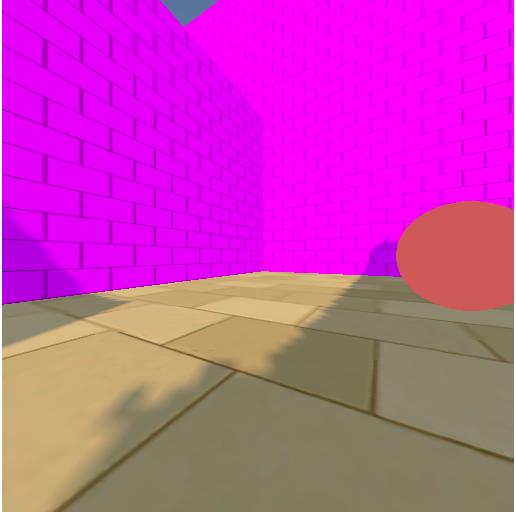}
    \captionof{figure}{Go(4);}
  \end{minipage}%
  \hfill
  \begin{minipage}[t]{0.3\textwidth}
    \centering
    \includegraphics[width=\linewidth]{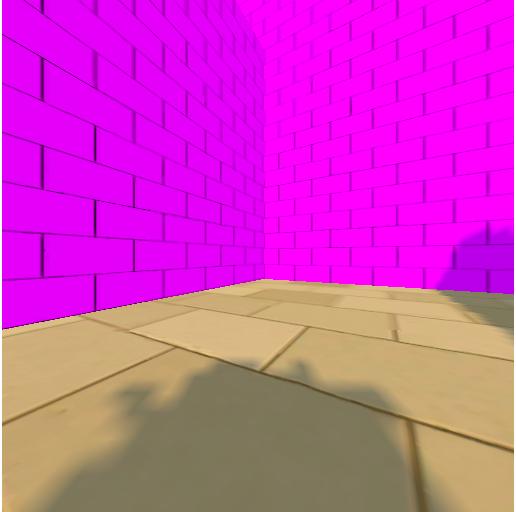}
    \captionof{figure}{Think(``The red ball is somewhere to my right: I'll make sure to get it in view before I continue");Turn(120);}
  \end{minipage}

    \begin{minipage}[t]{0.3\textwidth}
    \centering
    \includegraphics[width=\linewidth]{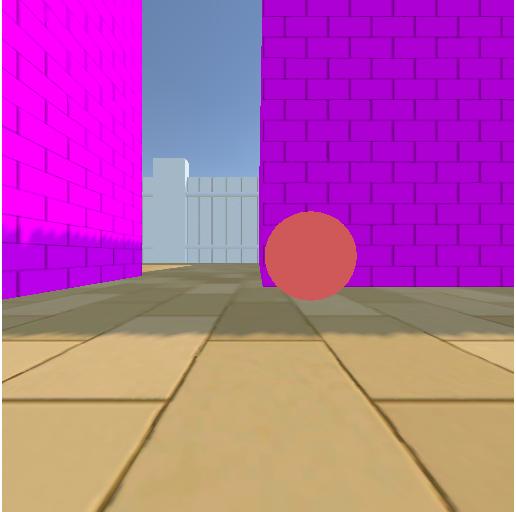}
    \captionof{figure}{Turn(-25);}
  \end{minipage}%
  \hfill
  \begin{minipage}[t]{0.3\textwidth}
    \centering
    \includegraphics[width=\linewidth]{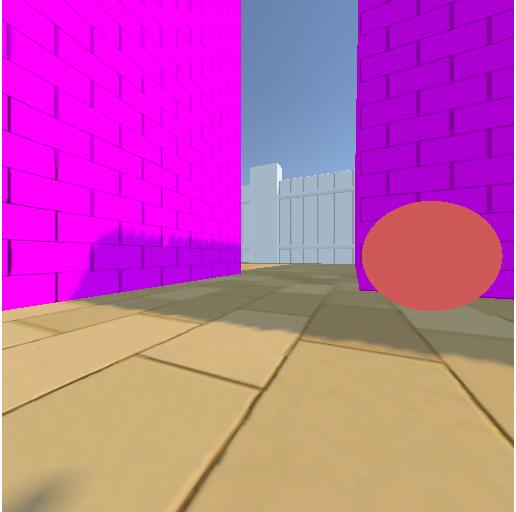}
    \captionof{figure}{Go(15);}
  \end{minipage}%
  \hfill
  \begin{minipage}[t]{0.3\textwidth}
    \centering
    \includegraphics[width=\linewidth]{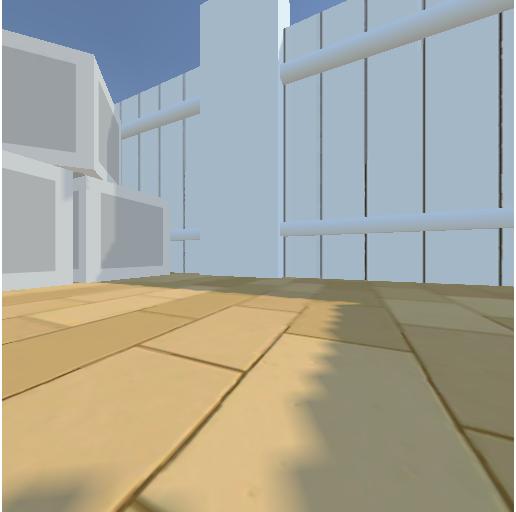}
    \captionof{figure}{Turn(-50);}
  \end{minipage}

    \begin{minipage}[t]{0.3\textwidth}
    \centering
    \includegraphics[width=\linewidth]{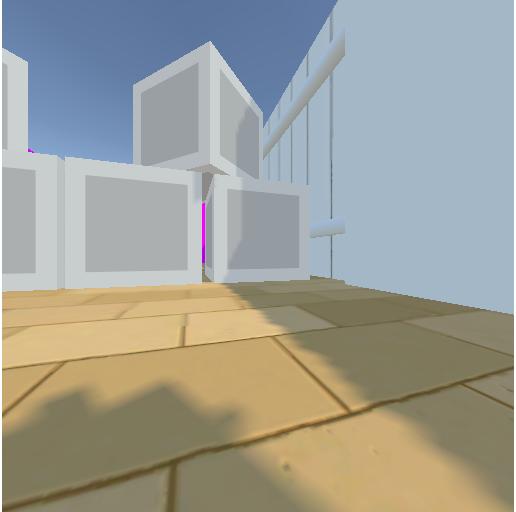}
    \captionof{figure}{Think(``There seem to be piled boxes in front of me: I'll push them out of the way by crashing into them with speed"); Go(15);}
  \end{minipage}%
  \hfill
  \begin{minipage}[t]{0.3\textwidth}
    \centering
    \includegraphics[width=\linewidth]{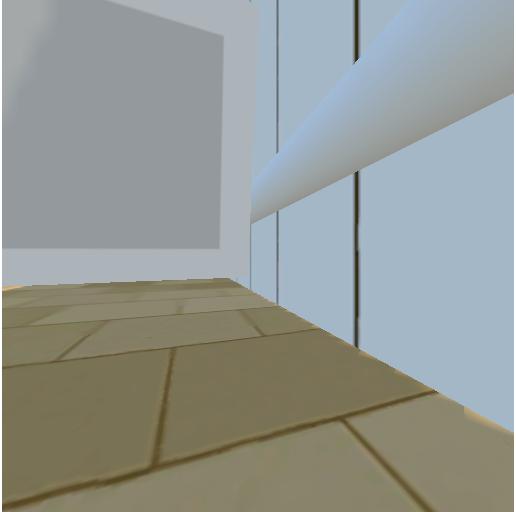}
    \captionof{figure}{Think(``I should confirm that I was successful in moving the boxes by turning around"); Turn(-120);}
  \end{minipage}%
  \hfill
  \begin{minipage}[t]{0.3\textwidth}
    \centering
    \includegraphics[width=\linewidth]{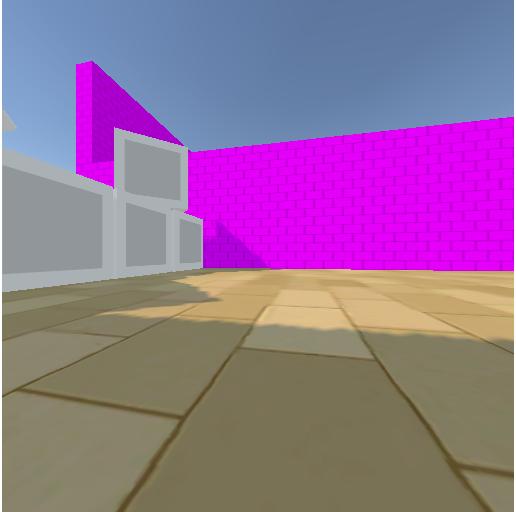}
    \captionof{figure}{Think(``I can see some piled boxes from the other side, so I have made it through. I'll turn to search this area for the reward"); Turn(90);}
  \end{minipage}

  \begin{minipage}[t]{0.3\textwidth}
    \centering
    \includegraphics[width=\linewidth]{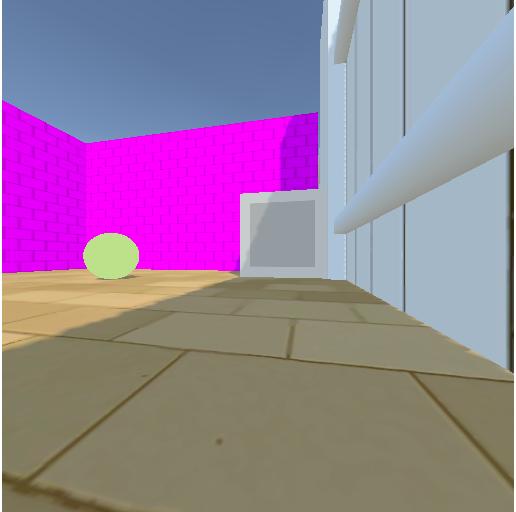}
    \captionof{figure}{Think(``The green ball is in view, I should turn about 30 degrees to my left to get it"); Turn(-30);}
  \end{minipage}%
  \hfill
  \begin{minipage}[t]{0.3\textwidth}
    \centering
    \includegraphics[width=\linewidth]{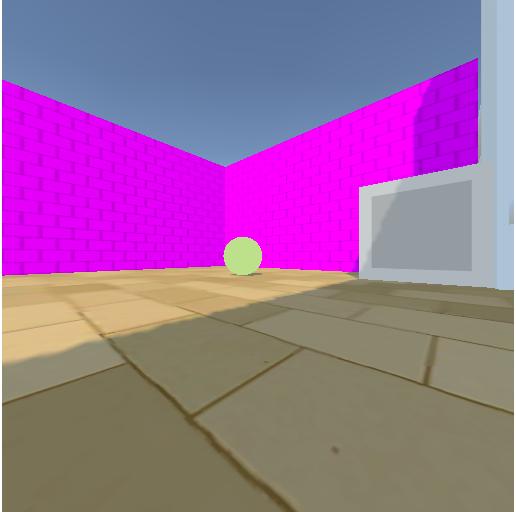}
    \captionof{figure}{Think(``The green ball is centered in my field of vision! I can advance forward to get it!"); Go(10);}
  \end{minipage}%
  \hfill
  \begin{minipage}[t]{0.3\textwidth}
  \end{minipage}
\end{center}

\end{document}